\documentclass[lettersize,journal]{IEEEtran}
\usepackage[compatibility=false]{caption}
\usepackage{amsmath,amsfonts}
\usepackage{algorithmic}
\usepackage{algorithm}
\usepackage{array}
\usepackage[caption=false,font=normalsize,labelfont=sf,textfont=sf]{subfig}
\usepackage{textcomp}
\usepackage{stfloats}
\usepackage{url}
\usepackage{verbatim}
\usepackage{graphicx}
\usepackage[numbers]{natbib}
\usepackage{cleveref}
\usepackage{color}
\usepackage{subcaption}
\usepackage{multirow}
\usepackage{booktabs}
\usepackage[table]{xcolor}
\usepackage{amssymb}

\usepackage{amsthm}
\theoremstyle{definition}
\newtheorem{definition}{Definition}
\newtheorem{theorem}{\rm\textbf{Theorem}}

\usepackage{xcolor}
\definecolor{codegreen}{rgb}{0, 0.6, 0}
\definecolor{codegray}{rgb}{0.5, 0.5, 0.5}
\definecolor{codepurple}{rgb}{0.58, 0, 0.82}
\definecolor{background}{rgb}{0.95, 0.95, 0.92}

\usepackage{etoolbox}
\newtoggle{final}

\begin{document}

\title{SSP: Safety-guaranteed Surgical Policy \\via Joint Optimization of Behavioral and Spatial Constraints}


\author{Jianshu Hu$^1$, Zhiyuan Guan$^1$, Lei Song$^2$, Kantaphat Leelakunwet$^1$,

Hesheng Wang$^1$, Wei Xiao$^3$, Qi Dou$^2$, Yutong Ban$^1$

\thanks{$^{1}$ Jianshu Hu, Zhiyuan Guan, Kantaphat Leelakunwet, HeSheng Wang and Yutong Ban are with Global College, Shanghai Jiao Tong University, Shanghai, China
        {\tt\small \{hjs1998, benjaminguan, kantaphat.lee, wanghesheng, yban\}@sjtu.edu.cn}}%
\thanks{$^{2}$Lei Song and Qi Dou are with the Department of Computer Science and Engineering, The Chinese University of Hong Kong,
        HongKong, China
        {\tt\small leisong, qidou@cuhk.edu.hk}}%
\thanks{$^{3}$Wei Xiao is with the Computer Science and Artificial Intelligence Laboratory, Massachusetts Institute of Technology,
        Massachusetts, America
        {\tt\small weixy@mit.edu}}%
\thanks{Corresponding email: {\tt\small  yban@sjtu.edu.cn}}%

}



\maketitle

\begin{abstract}
The paradigm of robot-assisted surgery is shifting toward data-driven autonomy, where policies learned via Reinforcement Learning (RL) or Imitation Learning (IL) enable the execution of complex tasks.
However, these ``black-box" policies often lack formal safety guarantees, a critical requirement for clinical deployment.
In this paper, we propose the Safety-guaranteed Surgical Policy (SSP) framework to bridge the gap between data-driven generality and formal safety.
We utilize Neural Ordinary Differential Equations (Neural ODEs) to learn an uncertainty-aware dynamics model from demonstration data.
This learned model underpins a robust Control Barrier Function (CBF) safety controller, which minimally alters the actions of a surgical policy to ensure strict safety under uncertainty.
Our controller enforces two constraint categories: behavioral constraints (restricting the task space of the agent) and spatial constraints (defining surgical no-go zones).
We instantiate the SSP framework with surgical policies derived from RL, IL and Control Lyapunov Functions (CLF).
Validation on in both the SurRoL simulation and da Vinci Research Kit (dVRK) demonstrates that our method achieves a near-zero constraint violation rate while maintaining high task success rates compared to unconstrained baselines.
\end{abstract}


\section{Introduction}




\IEEEPARstart{T}{he} field of medical robotics stands at a precipice of transformation.
For the past two decades, the da Vinci Surgical System and its contemporaries have operated primarily under a teleoperation scheme, amplifying the dexterity but relying fully on human supervision. Recently, the development of simulation platforms, datasets and data-driven robot learning methods are propelling the field toward surgical autonomy, where robots autonomously execute subtasks such as suturing, debridement, and material handling.
In particular, learning-based methods, such as Deep Reinforcement Learning (DRL) and Imitation Learning (IL), have demonstrated impressive capabilities in acquiring dexterous surgical skills from data \citep{SRT-H, doi:10.1126/scirobotics.adt3093}.
However, such data-driven methods often lack formal safety guarantees, which are crucial in surgical applications.

Ensuring the safety and efficacy of robot-assisted surgical manipulations requires the simultaneous satisfaction of two critical, yet often competing, objectives: (i) precise reference path following, and (ii) strict no-go zone avoidance.
To achieve the desired clinical outcome, the robotic manipulator must accurately track a prescribed reference path that represents the surgical intent, maintaining high precision despite unmodeled dynamics or environmental disturbances.
Concurrently, the surgical field is highly constrained and densely populated with vital anatomical structures, such as major blood vessels and nerve bundles that define absolute forbidden regions.
Consequently, a fundamental challenge in designing surgical control architectures lies in hierarchically prioritizing these requirements, guaranteeing that the pursuit of tracking accuracy never compromises the integrity of these surrounding ``no-go zones'' \cite{laplante2023validation} or leads to irreversible injury.
Moreover, since training datasets cannot exhaustively cover every possible variation or corner case, black-box policies are prone to unpredictable behaviors in unseen scenarios, potentially leading to catastrophic outcomes such as damaging vital organs or tearing tissue.

On the other hand, purely rule-based or classical control methods, while capable of providing rigorous mathematical guarantees for safety, often suffer from notably low performance in complex surgical tasks. Since these traditional approaches rely heavily on explicit, hand-crafted analytical models, they frequently struggle to adapt to the highly non-linear, deformable nature of soft tissues and unexpected environmental variations. To maintain strict safety margins without the benefit of learned adaptability, rule-based systems are typically forced into overly conservative behaviors. This rigidity can result in a low success rate in many surgical manipulation tasks, thereby highlighting the critical need for a hybrid framework that marries the dynamic performance of learned policies with the absolute safety guarantees.

\begin{figure}[t]
    \centering
    \includegraphics[width=\linewidth]{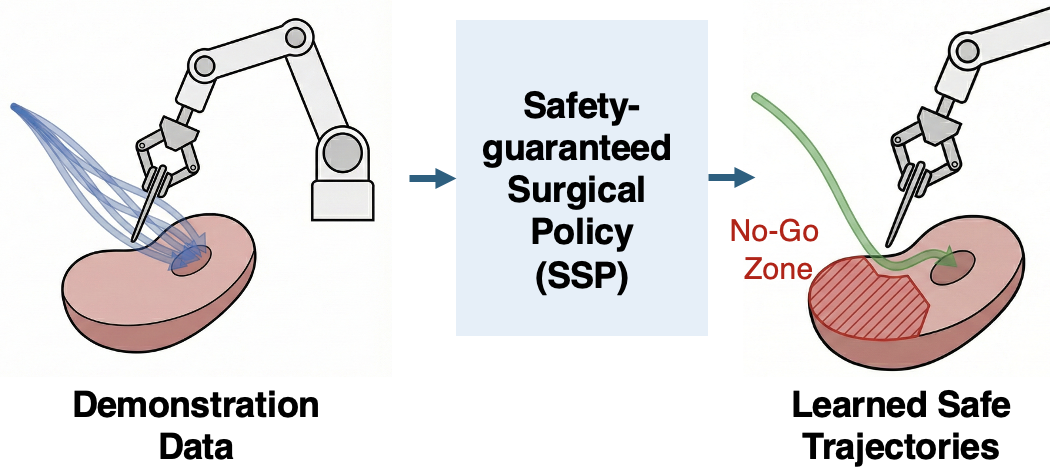} 
    \caption{\textbf{Safety-guaranteed Surgical Policy Framework:} We propose a safety-guaranteed surgical policy framework, which learns robust and safe executions of surgical actions.}
\end{figure}

To address this, we propose a Safety-guaranteed Surgical Policy (SSP) framework that decouples task performance from safety assurance.
A surgical policy, which can be any black-box policy pre-trained via RL/IL or a path following controller defined by Control Lyapunov Functions (CLF), provides a nominal action focused on task completion.
We introduce a Robust Control Barrier Function (CBF) safety controller that acts as a rigorous "safety filter", overriding the surgical policy when a violation of the safe set is imminent.
Unlike heuristic safety checks, CBFs provide a mathematical framework for set invariance, ensuring that if the system starts in a safe state, it remains there indefinitely.
This whole architecture allows the robot to exploit the adaptability of learning-based policies while adhering to the rigorous safety bounds defined by the surgical context.
Since CBF relies on instantaneous time derivatives to guarantee safety, we propose employing \textit{Neural Ordinary Differential Equations} (Neural ODEs) to learn an uncertainty-aware dynamics model. To prevent the safety filter from dangerously hallucinating safety in regions with high epistemic uncertainty, we integrate the prediction error of the Neural ODE directly into the barrier formulation to adaptively contract the safe set. Furthermore, we introduce an additional behavioral CBF in task space that confines the agent to the valid training distribution of the learned dynamics, preventing the agent from drifting into Out of Distribution (OOD) states where the dynamics are unknown and thereby preserving the absolute integrity of the theoretical safety guarantees throughout the surgical manipulation.

The main contributions are as follows:
\begin{itemize}
    \item We propose a unified Safety-guaranteed Surgical Policy framework that integrates Neural ODEs for uncertainty-aware dynamics learning, demonstration-guided policy generation and CLF-based path following for surgical tasks, and robust CBFs for safety requirement. This structure allows for safe deployment of ``black-box" policies by wrapping them in a theoretically guaranteed safety controller.

\item  We formulate a robust CBF-QP that incorporates a quantified uncertainty term. We consider a novel Behavioral control barrier function (Behavioral CBF) that constrains the agent to be close to the demonstration data distribution and a Spatial control barrier function (Spatial CBF) restricts the agent to stay away from no-go zone. The incorporation of these three terms ensure robust safety during deployment of the surgical policy.

\item We provide extensive empirical validation in both SurRoL simulation environment and on a real-world dVRK.
We demonstrate that our method achieves a low constraint violation rate while maintaining high task success rates compared to unconstrained baselines.
And we analyze the performance of our framework under different tasks and different constraints.
\end{itemize}

\IEEEpubidadjcol 
\section{Related Work}
In this section, we introduce the related work about policy learning for surgical robot, Neural ODE for dynamics modelling and CBF for safety guarantee.

\subsection{Policy Learning for Surgical Robot}
The field of robot learning has witnessed substantial advances, driven primarily by the scaling of data and network architectures.
In surgical robotics, a impactful trend is the application of learning-based methods \citep{doi:10.1126/scirobotics.adt3093}, where reinforcement learning (RL) and imitation learning (IL) are the predominant paradigms.
To address the sample inefficiency of pure RL in contact-rich environments, researchers have successfully integrated expert demonstrations into the training loop.
The DEX \cite{dex} algorithm utilizes a non-parametric regression of the expert actions to guide exploration, significantly accelerating learning for tasks like needle picking.
Furthermore, this paradigm has been scaled to long-horizon tasks through skill-chaining \citep{Vi-skill}.

More recently, the concept of "Surgical Embodied Intelligience" \citep{doi:10.1126/scirobotics.adt3093} has expanded the scope of autonomy to generalized tasks using large-scale datasets.
Similarly, SRT-H \citep{SRT-H} introduces a hierarchical framework conditioned on language instructions, allowing surgeons to command high-level sub-tasks.
With a large amont of high-quality demonstration data, SRT-H demonstrates perfect success rates in challenging real surgical environments.
Despite their proficiency, these methods often lack specific consideration on the critical safety constraints required for real-world surgical applications.
Our framework complements these learning-based approaches by providing necessary safety guarantees.

\subsection{Neural ODE for Dynamics Modeling}
Effective model-based control requires an accurate representation of system dynamics.
In surgical environments, analytical derivation of interaction dynamics (e.g., needle-tissue friction, cutting forces) is notoriously difficult and computationally expensive.
An alternative powerful method for learning dynamics is to learn from data.
Neural Ordinary Differential Equations (Neural ODEs) \citep{chen2019neuralordinarydifferentialequations}, which parameterize the continuous-time derivative of the state $\dot s$, have been successfully applied to problems like system identification \citep{duong2024porthamiltonianneuralodenetworks} \citep{kim2021stiff} \citep{rahman2022neural} and model-based control \citep{kasaei2023data}.
In robotics, this continuous formulation is crucial for computing the Lie derivatives required for barrier functions \citep{xiao2023inv}.
However, few studies have considered uncertainty quantification for neural ODEs, which is crucial for strictly enforcing the system performance such as safety.
In this work, we leverage the capability of Neural ODEs to accurately learn the continuous-time dynamics of the daVinci robot, while performing uncertainty quantification for the learned neural ODE.
We also define a task space CBF to ensure the system state staying within the task space where the neural ODE is trained in order to ensure the reliability of the model. 

\subsection{CBF for Safety Guarantee}
Control Barrier Functions (CBFs) \citep{7782377} \citep{Glotfelter2017} \citep{xiao2021high} have emerged as a primary tool for enforcing set invariance in safety-critical systems.
The CBF method can map a nonlinear state constraint onto another constraint that is linear in control, and the satisfaction of the control constraint implies the satisfaction of the original nonlinear state constraint.
In such a way, the CBF method can transform a nonlinear optimization problem into a quadratic program (QP) that minimizes the deviation from a nominal control input subject to CBF constraints \citep{7782377}.
CBFs provide a safety filter that is active only when necessary, and they are widely applied in robotic tasks such as safe navigation \citep{9718195} \citep{gonzalez2025safe}, robot swarm formation control \citep{ceron2024reciprocal}, robot manipulation \citep{tang2024learning} \citep{brunke2025semantically} and safe robot learning \citep{xiao2023safediffuser} \citep{xiao2023barriernet}.

However, traditional CBF formulations assume perfect knowledge of the system dynamics.
A recent and active area of research is the development of CBFs for systems with learned dynamics, such as those approximated by neural networks \citep{xiao2023inv}.
This line of work bridges the gap between data-driven modeling and formal safety guarantees.
However, there is still a mismatch between the learned dynamics and real model.
In this work, we combine the CBF, neural ODEs and uncertainty quantification in the same framework to strictly guarantee the safety of the system with the learned model.



\section{Background}
In this section, we formalize the mathematical foundations of our framework, covering dynamics modeling with neural ODE and basics of CBF and CLF.

\subsection{Neural Ordinary Differential Equations}
Instead of modeling the discrete state transition $s_{t+1} = f(s_t, a_t)$, Neural ODEs model the continuous-time derivative of the state.
We define the system dynamics as a control-affine ODE parameterized by a neural network $\eta$:
\begin{equation}
    \dot{s}(t) = f_\eta(s(t))+g_{\eta}(s(t)) a(t),
\end{equation}
where $s(t) \in \mathcal{S}$ is the state and $a(t) \in \mathcal{A}$ is the control input.
$f_{\eta}$ represents the drift dynamics and $g_\eta$ is the control matrix.
The state at any future time $\Delta t$ is computed by integrating this ODE starting from an initial state $s_{t_0}$:
\begin{equation}
\label{eq:neural ode integral}
    s(t_0+\Delta t) = s(t_0) + \int_{t}^{t+\Delta t} \left( f_\eta(s(\tau)) + g_\eta(s(\tau))a(\tau) \right) d\tau.
\end{equation}
This integral is solved using a numerical ODE solver.
The continuous-time formulation is essential for our safety framework because CBFs rely on the time derivative of the barrier function.

\subsection{Control Barrier Function and Control Lyapunov Function}
\begin{definition}[Control Barrier Function \citep{7782377}]
Let $b(s):\mathbb{R}^n \rightarrow \mathbb{R}$ be a continuously differentiable function.
If it satisfies:
\begin{equation}
\begin{aligned}
    &\mathcal{C}:= \{b(s)>0\},\\
    &\exists a, L_fb(s)+L_gb(s)a+ \gamma b(s)\geq 0, \forall s \in \mathcal{C},
\end{aligned}
\end{equation}
we say $b(s)$ is a control Barrier function and $\mathcal{C}$ is an invariant set.
$L_fb(s) = \nabla b(s)^T f(s)$ and $L_gb(s) = \nabla b(s)^T g(s)$ are the Lie derivatives of $b$ along the vector fileds $f$ and $g$.
\end{definition}
The condition implies that as the system approaches the boundary of the safe set, there is always a control input to ensure $\dot b(s)$ to be positive (or not too negative) to prevent the system from leaving $\mathcal{C}$.

\begin{definition}[Control Lyapunov Function \citep{ames2012control}]
Let $V(s):\mathbb{R}^n \rightarrow \mathbb{R}$ be a continuously differentiable function.
If it satisfies:
\begin{equation}
\begin{aligned}
    &\Omega_c:= \{\mathbf{s}\in \mathbb{R}^n|V(s)\leq c, c>0\}\\
    &V(s)>0, \forall s \neq \mathbf{0}, V(\mathbf{0})=0\\
    &\exists a, L_fV(s)+L_gV(s)a+\beta V(s) \leq 0, \forall s \in \Omega_c \backslash \{\mathbf{0}\}
\end{aligned}
\end{equation}
we say $V(s)$ is a control Lyapunov function and $\Omega_c$ is an invariant set.
\end{definition}

This condition ensures that there exists a control input that decreases the Lyapunov function $V(s)$ over time, leading the system toward the goal state where $V(s) = 0$.

\section{Methodology}
\begin{figure*}[tb]
    \centering
    \includegraphics[width=\linewidth]{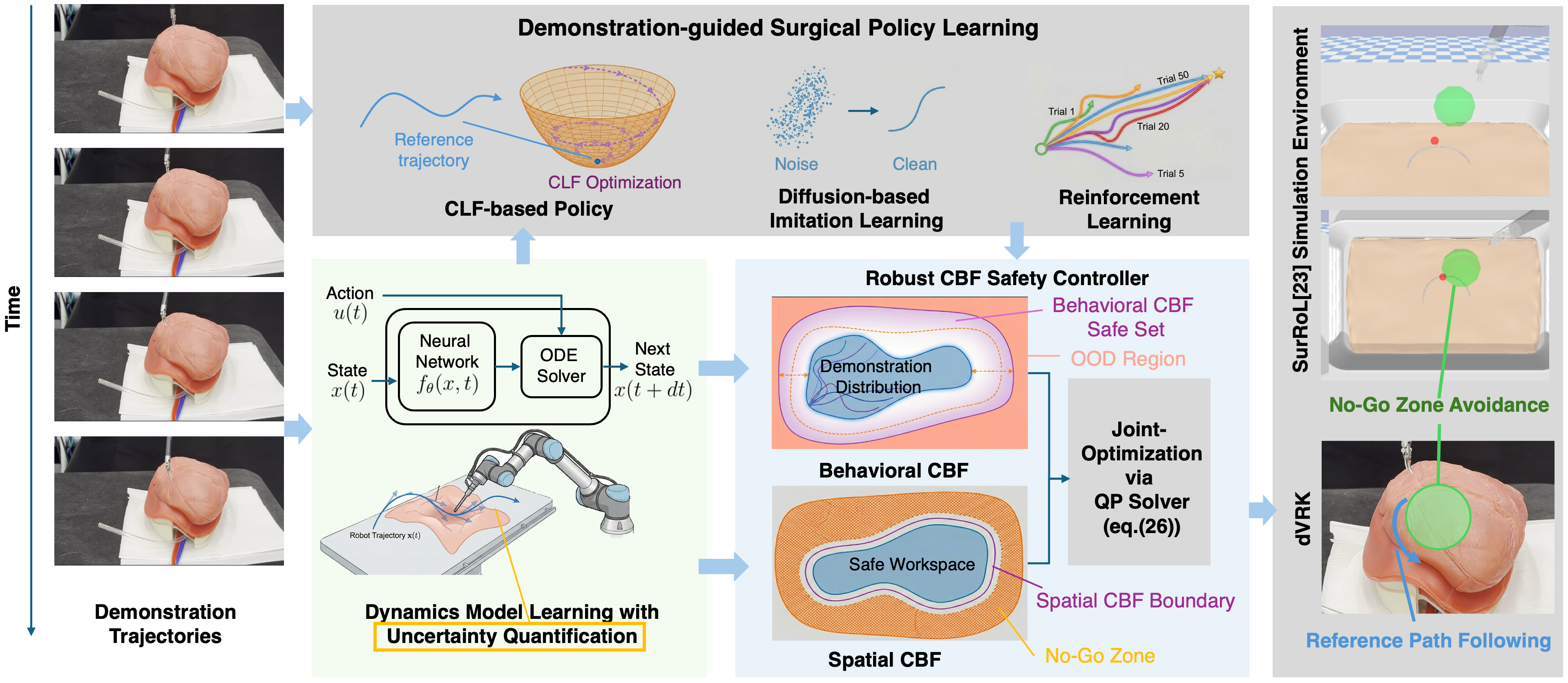} 
    \caption{\textbf{Overview of the Safety-guaranteed Surgical Policy (SSP) Framework:} This architecture decouples task performance from safety assurance by wrapping ``black-box" surgical policies within a theoretically guaranteed safety layer.
    The framework utilizes Neural Ordinary Differential Equations (Neural ODEs) to learn a continuous dynamics model with uncertainty quantification, which underpins a Robust Control Barrier Function (CBF) controller.
    By solving a quadratic program that jointly optimizes for behavioral constraints (restricting the agent to the valid task space) and spatial constraints (avoiding no-go zones), the system minimally deviates from nominal actions to ensure strict safety during deployment.}
    \label{fig:overview}
\end{figure*}
We present our Safety-guaranteed Surgical Policy (SSP) framework designed for safety-guaranteed surgical manipulations.
The framework consists of three integrated modules:
a) Continuous Dynamics Model Learning with Uncertainty Quantification.
b) Demonstration-guided Surgical Policy Learning c) Robust CBF Safety Controller.
This framework ensures that the robot exploits the adaptability of learning-based methods for complex manipulation while adhering to rigorous safety bounds.


\subsection{Safety-guaranteed Surgical Policy Framework}
We propose a Safe-guaranteed Surgical Policy (SSP) framework that decouples task performance from safety assurance, enabling the safe deployment of ``black-box" policies in safety-critical surgical environments.
Formally, we consider a robotic system with state $s \in \mathcal{S}$ and control input $a \in \mathcal{A}$.
The objective is to perform a surgical task specified by a surgical policy $\pi_{task}$ while strictly satisfying a set of safety constraints defined by a safe set $\mathcal{C}$.
As illustrated in \Cref{fig:overview}, the framework consists of three interconnected modules: 

Firstly, the framework includes a dynamics model learning module with uncertainty quantification. Since analytical models of surgical environments are often imprecise due to complex interactions, we first approximate the continuous-time evolution of the system $\dot{s} = f(s, a)$ using a Neural Ordinary Differential Equation (Neural ODE).
Moreover, we consider an uncertainty term $\epsilon$ in the system dynamics $\dot{s} = f(s, a)+ \epsilon$.
This learned model provides the gradient information necessary for derivative-based safety constraints.
To quantify the uncertainty in the learned model, we define a task space ($\mathcal{T} \subset \mathcal{S}$) where the agent needs to stay for reliable Neural ODE prediction. 

Secondly, a demonstration-guided surgical policy learning module is required. A guiding policy $\pi$ generates nominal actions $a_{des}$ aimed at solving the task.
Our framework is agnostic to the policy type. It can be a learning-based policy obtained from reinforcement learning or imitation learning, or a path follower based on Control Lyapunov Functions (CLF).

Finally, the framework includes a Robust CBF Safety Controller. The nominal action $a_{des}$ serves as the reference for a safety controller based on Control Barrier Functions (CBF).
This controller solves a real-time optimization problem that minimally deviates from $a_{des}$ to synthesize a safe control input $a_{safe}$.
Here, we consider two types of constraints the agent needs to satisfy:
a) behavioral constraint which is defined by the task space b) spatial constraint which is defined by the no-go zone of the surgical task.


\subsection{Dynamics Model Learning with Uncertainty Quantification}
Effective model-based control requires an accurate representation of the system dynamics.
In surgical environments, such as the dVRK environments or SurRoL \citep{xu2021surrol}, the state $s \in \mathcal{S}$ is defined by the position of the end-effector $x \in \mathbb{R}^3$ and orientation $\theta \in \mathbb{R}^3$, denoted as $s = [x, \theta]$.
The action $a \in \mathcal{A}$ consists of the linear velocity control $u$ and angular velocity control $u_{\theta}$, such that $a = [u, u_{\theta}]$.

We approximate the continuous-time dynamics of the robot using Neural Ordinary Differential Equations (Neural ODEs). 
We model the system as a control-affine system: 
\begin{equation}
    \dot{s} = f_{\eta}(s)+g_{\eta}(s)a + \epsilon,
\label{eq:neural ode for whole state}
\end{equation}
where $f_{\eta}$ and $g_{\eta}$ represent the learned dynamics and control input matrix, which are parameterized by $\eta$, and $\epsilon$ is the uncertainty term to be verified and quantified during inference time.
The network is trained on a demonstration dataset $\mathcal{D}$ to minimize the integration error over a time horizon.
Taking a state-action sequence $d = (s_0,a_0,s_1,a_1,...,s_h)$ from the dataset $\mathcal{D}$, we calculate the integral using \Cref{eq:neural ode for whole state} to get the predicted state sequence $(\hat{s}_1,...,\hat{s}_h)$ and the loss:
\begin{equation}
\label{eq:neural_ode_loss}
    L = \sum_{i=1}^{h} \|s_i-\hat{s}_i\|_1.
\end{equation}
For the specific purpose of positional no-go zone, we also learn the another Neural ODE which only considers the position of the end effector:
\begin{equation}
    \dot{x} = f_{\psi}(x)+g_{\psi}(x)u  + \epsilon,
\label{eq:neural ode for position}
\end{equation}
where the network is parameterized by $\psi$. 


To rigorously quantify the uncertainty term $\epsilon$ for our safety filter, we calculate two distinct error metrics.
Consider a transition $(s_t, a_t, s_{t+1})$ we obtain at time step $t$ during policy execution.
Firstly, we measure the derivative prediction error ($E_{\dot s}$), which captures the instantaneous dynamics mismatch.
This compares the predicted time derivative against the true state derivative $\dot{s}^*$ (computed via finite difference ):
\begin{equation}
\label{eqn:un1}
\begin{aligned}
    E_{\dot s} &= \max_{t} \| \dot{s}_{t}^*- \dot{s}_{t} \|_1,\\
    \dot{s}_{t} &= f_{\eta}(s_t) + g_{\eta}(s_t)a_t.
\end{aligned}
\end{equation}

Secondly, we measure the state prediction error ($E_{s}$), which evaluates the accuracy on the integration of the Neural ODE. 
For each step $t$, we integrate the Neural ODE starting from the ground truth current state $s_t$ to predict the next state $\hat{s}_{t+1}$.
We define the metric as the maximum prediction error observed within a trajectory:
\begin{equation} \label{eqn:un2}
\begin{aligned}
    E_{s} &= \max_{t} \| s_{t+1} - \hat{s}_{t+1}\|_1,\\
    \hat{s}_{t+1} &= s_t + \int_{t}^{t+\Delta t} (f_{\eta}(s(\tau)) + g_{\eta}(s(\tau))a_t) d\tau.
\end{aligned}
\end{equation}

Finally, to account for the uncertainty inherent in the learned dynamics, we formally define a valid task space $\mathcal{T} \subset \mathcal{S}$.
Intuitively, since the Neural ODE is trained on a finite set of demonstrations $\mathcal{D}$, the prediction error $\epsilon$ can become significant in out-of-distribution (OOD) regions, making the learned derivatives $f_{\eta}$ and $g_{\eta}$ unreliable.
If the agent drifted into these regions, the Lie derivatives calculated for the CBF optimization would be inaccurate, potentially leading to unreasonable or unstable control updates.
To prevent this, we treat the boundary of $\mathcal{T}$ as a hard safety constraint.
By strictly confining the agent within this region, we ensure that the robust CBF safety controller always operates with a dynamics model of high accuracy, guaranteeing that the generated safe actions are physically consistent and effective.
We define the task space around the demonstration states:
\begin{equation}
\label{eq:task space}
    \mathcal{T}:\{s|\min_{s_D \in D} \|s-s_D\|_2\leq d, s\in \mathcal{S}\}.
\end{equation}
Correspondingly, we define the barrier function $b_{\mathcal{T}}(s)$ for this task space:
\begin{equation}
\begin{aligned}
\label{eq:task space barrier function}
    b_{\mathcal{T}}(s) &= d^2-\|s-s_{min}\|_2^2,\\ s_{min} &= argmin_{s_D\in D} \|s-s_D\|_2.
\end{aligned}
\end{equation}

\subsection{Demonstration-guided Surgical Policy Learning}
We consider three typical types of methods to generate the nominal action $a_{des}$: a demonstration-guided policy trained with RL, a diffusion-based policy learned by imitation learning and a path following policy defined by the Control Lyapunov Function.

\noindent \textbf{Demonstration-guided RL}
For tasks with rewards which can not be easily defined, we can consider using a demonstration-guided RL policy $\pi_\phi$.
We utilize the DEX \citep{dex}, which augments the Deep Deterministic Policy Gradient (DDPG) \citep{lillicrap2019continuouscontroldeepreinforcement} framework with expert demonstrations to guide exploration in sparse-reward surgical environments.
DEX incorporates an additional reward for penalizing the gap between the agent policy and the expert policy:
\begin{equation}
    r = -D(a_t, a_t^e), a_t=\pi(s_t), a_t^e = \pi^e(s_t),
\end{equation}
where $D$ is a distance metric used to measure the gap between the agent action $a_t$ and the expert action $a_t^e$ at time step $t$. 
Instead of learning a policy parameterized by a neural network from the demonstration data, DEX uses non-parametric regression model:
\begin{equation}
    \pi^e = \frac{\sum_{i=1}^N exp(-\|s-s^i\|_2)\cdot a^i}{\sum_{i=1}^N exp(-\|s-s^i\|_2)},
\end{equation}
where $\{s^i|i=1,..,N\}$ are the $N$ nearest neighbors of the state $s$ within a minibatch from the expert demonstrations and $\{a^i|i=1,...,N\}$ are the corresponding expert actions. 
Intuitively, this model assumes similar states share similar optimal actions.

\noindent \textbf{Diffusion-based Imitation Learning}
Given a set of demonstrations, we train a diffusion policy \citep{chi2023diffusionpolicy} $\pi_\phi(a|s)$ conditioned on the states $s$, by learning the conditional score function of the data distribution.
The training and inference of a diffusion policy contains a diffusion process and a denoising process.
In the diffusion process, scheduled Gaussian noise with variance $\mu^k$ is gradually added to the clean action $a^0$ at diffusion step $k$:
\begin{equation}
    q(a^k|a^{k-1}) = \mathcal{N}(a^k;\sqrt{1-\mu^k}a^{k-1},\mu^k\mathbf{I}).
\end{equation}
To avoid confusion, we use the superscript $k$ to indicate the diffusion step, which is different from the subscript $t$ indicating the time step in a trajectory.
With the noisy actions $a^k$, the diffusion model is trained to predict the noise added to it given the diffusion step $k$ and the states $s$. 
The following loss function is used to train the diffusion model:
\begin{equation}
\mathcal{L} = \mathbb{E}_{s, a\sim \mathcal{D}} \Big[ \mathbb{E}_{a^0, a^k}||\epsilon^k-g_\phi(a^k,k,s)||^2 \Big],
\end{equation}
where $(s,a)$ are state-action pairs sampled from the demonstration dataset $\mathcal{D}$, $a^k$ is the noisy action, $\epsilon^k$ is the noise added at diffusion step $k$, and $g_\phi$ is the diffusion model.

During inference, to sample action from the diffusion policy $\pi_\phi(a|s)$, we need to first sample from a Gaussian distribution to get a noisy action $a^k$, and then repeat the denoising step with the learned score function $g_\phi$:
\begin{equation}
    a^{k-1} = \alpha_1 (a^k - \alpha_2 g_\phi(a^k, k, s) ) + \mathcal{N}(0, \alpha_3 \mathbf{I}),
\end{equation}
where $\alpha_1$, $\alpha_2$, $\alpha_3$ are all constant related to the noise scheduler, only depending on $k$, used in the diffusion process.

\noindent \textbf{CLF-based Policy}
For tasks defined by a reference path $\rho^* = [s^*_0, s^*_1, s^*_2, ..., s^*_M]$, such as a straight line for tissue cutting, we formulate a tracking controller based on a Control Lyapunov Function (CLF).
We define a quadratic Lyapunov candidate function $V(s)$ based on the deviation from a desired state $s_{des}$:
\begin{equation}
    V(s) = \|c(s-s_{des})\|^2,
    \label{eq:lyapunov_function}
\end{equation}
where $c$ is a positive constant.
The stability condition requires that the time derivative of $V$ decreases exponentially:
\begin{equation}
    \dot{V}(s) = L_f V(s) + L_g V(s)a \leq -\beta V(s),
\end{equation}
where $\beta$ is a hyperparameter for controlling the optimization problem, $L_f V$ and $L_g V$ are the Lie derivatives of $V$ along the dynamics learned in \Cref{eq:neural ode for whole state}.
Formally, the optimization problem is defined as
\begin{equation}
\begin{aligned}
        \min &\quad \|a\|^2 \\
        \text{s.t. } &\dot{V}(s)+\beta V(s) \leq 0.
\end{aligned}
\label{eq:clf optimization problem}
\end{equation}

The problem now is to decide the desired state $s_{des}$, which are needed for defining the optimization problem, considering the predefined path $\rho^*$.
First, we find the closest state in the pre-defined trajectory $\rho^*$ given current state $s$. 
Assume the closest state in the pre-defined trajectory is $s^*_i$ at index $i$, to make sure the agent is moving forward along the trajectory, we first set $s^*_i$ as the desired state and move to next state $s^*_{i+1}$ until the distance between current state $s$ and desired state $s^*_c$ is smaller than a threshold $\Delta$:
\begin{equation}
    \|s-s^*_i\|^2 < \Delta.
\end{equation}

\begin{algorithm}[t]
\caption{SSP: Safety-guaranteed Surgical Policy}
\label{alg:ssp}
\begin{algorithmic}[1]
\REQUIRE Demonstration dataset $\mathcal{D}$, Safety set $\mathcal{C}$
\STATE \textbf{Phase 1: Dynamics and Policy Learning}
\STATE Define Task Space $\mathcal{T}$ (\Cref{eq:task space}) and barrier $b_{\mathcal{T}}(s)$ based on $\mathcal{D}$ (\Cref{eq:task space barrier function}).
\STATE Train Neural ODE parameters $\eta$ by minimizing $L_1$ loss (\Cref{eq:neural_ode_loss}) on $\mathcal{D}$
\STATE Learn a surgical policy $\pi_{task}$ from $\mathcal{D}$ or define a reference path $\rho^*$ for the surgical task.
\STATE Quantify uncertainty metrics $E_{\dot{s}}$ and $E_{s}$ using \Cref{eqn:un1} and \Cref{eqn:un2}.
\STATE \textbf{Phase 2: Online Execution}
\WHILE{Task not completed}
    \STATE Observe current state $s_t$
    \STATE Generate nominal action $a_{des} \leftarrow \pi_{task}(s_t)$ (RL, IL, or CLF)
    \STATE Calculate uncertainty-aware set $Y(s_t) = \{y \mid s_t - E_s \le y \le s_t + E_s\}$
    \STATE \textbf{Safety Controller:}
    \STATE Solve Robust CBF-QP (\Cref{eq:cbf optimization problem}) 
    \STATE Execute $a_{safe}$ on surgical robot
\ENDWHILE
\end{algorithmic}
\end{algorithm}

\subsection{Robust CBF Safety Controller}
\label{method:cbf}
Regardless of whether the guiding action $a_{des}$ is generated by RL, IL or CLF, we apply a final robust CBF safety controller to enforce surgical constraints, such as spatial constraint (no-go zone avoidance), and behavioral constraints (remaining in valid task space).
We explain details of how we consider uncertainty of the Neural ODE to ensure safety under uncertainty.

\textbf{Spatial CBF.} The safety requirement is enforced by ensuring the forward invariance of $\mathcal{C}$, which leads to the condition:
\begin{equation} \label{eqn:cbf1}
    \dot{b}(s)\geq -\gamma b(s),
\end{equation}
where $\gamma$ is a positive constant.
Expanding $\dot{b}(s)$ using the dynamics from \Cref{eq:neural ode for whole state}, we obtain the constraint on the control input:
\begin{equation}
    L_fb(s)+L_gb(s)a + \frac{db(s)}{ds}\epsilon+\gamma b(s) \geq 0.
\end{equation}
Since $\epsilon$ in the above is general unknown, we replace the above CBF constraint by a robust CBF constraint in the form:
\begin{equation}
    L_fb(s)+L_gb(s)a - \left|\frac{db(s)}{ds}\right| E_{\dot s}+\gamma b(s) \geq 0.
\end{equation}

The state $s$ also introduces some uncertainties for the above robust CBF using the learned neural ODEs or under observation noise.
We further define the set of state $Y(s)$ from uncertainty quantification:
\begin{equation}
    Y(s) = \{y|s-E_{s} \leq y\leq s+ E_{s}\}.
\end{equation}

Finally, we define a robust CBF considering both state and dynamics uncertainties:
\begin{equation} \label{eqn:cbfn}
    \min_{y \in Y(s)}\left[L_fb(y)+L_gb(y)a - \left|\frac{db(y)}{dy}\right| E_{\dot s}+\gamma b(y)\right] \geq 0.
\end{equation}

Formally, the CBF optimization problem is defined as

\begin{align}
    \min &\quad \|a-a_{des}\|^2 \label{eq:cbf optimization problem} \\
    \text{s.t. } &\min_{y \in Y(s)}\left[L_fb(y)+L_gb(y)a - \left|\frac{db(y)}{dy}\right| E_{\dot s}+\gamma b(y)\right] \geq 0, \nonumber \\
    &\min_{y \in Y(s)}\left[L_fb_{\mathcal{T}}(y)\!+\!L_gb_{\mathcal{T}}(y)a \!-\! \left|\frac{db(y)}{dy}\right| E_{\dot s}\!+\!\gamma b_{\mathcal{T}}(y)\right] \geq 0 \nonumber
\end{align}

The second robust CBF, corresponding to the behavioral constraint $b_{\mathcal{T}}(y)$, is derived similarly as the one of spatial constraint $b(s)$ through (\ref{eqn:cbf1})-(\ref{eqn:cbfn}), namely \textbf{behavioral CBF}.

We have the following theorem to show the safety of the controller (\ref{eq:cbf optimization problem}):
\begin{theorem}
    If the robot is initially safe, then the CBF-based controller (\ref{eq:cbf optimization problem}) ensures the safety of the robot with the learned neural ODE model (\ref{eq:neural ode for whole state}) and the corresponding uncertainty quantification (\ref{eqn:un1}) (\ref{eqn:un2}).
\end{theorem}
\textit{Proof:} The second behavioral robust CBF constraint in (\ref{eq:cbf optimization problem}) ensures that the real robot state belongs to $Y(s)$ and the uncertainty $\epsilon$ in the neural ODE (\ref{eq:neural ode for whole state}) stays within the bound defined by $E_{\dot s}$ in (\ref{eqn:un1}) (i.e., $|\epsilon|\leq E_{\dot s}$).

Since $|\epsilon|\leq E_{\dot s}$, we have that 

\begin{align}
    &L_fb(s)+L_gb(s)a + \frac{db(s)}{ds}\epsilon+\gamma b(s) \\
    &\geq L_fb(s)+L_gb(s)a - \left|\frac{db(s)}{ds}\right| E_{\dot s}+\gamma b(s) \nonumber \\
    &\geq \min_{y \in Y(s)}\left[L_fb(y)+L_gb(y)a - \left|\frac{db(y)}{dy}\right| E_{\dot s}+\gamma b(y)\right] \geq 0. \nonumber 
\end{align}

The $L_fb(s)+L_gb(s)a + \frac{db(s)}{ds}\epsilon+\gamma b(s)$ is equivalent to $\dot b(s)+\gamma b(s)$. Therefore, following the last equation, we have
\begin{equation}
    \dot b(s)+\gamma b(s) \geq 0.
\end{equation}
By the CBF theorem \citep{7782377}, the last equation implies that $b(s)\geq 0$ if the robot is initially safe. Thus, we conclude that the CBF-based controller (\ref{eq:cbf optimization problem}) ensures the safety of the robot with the learned neural ODE model (\ref{eq:neural ode for whole state}) and the corresponding uncertainty quantification (\ref{eqn:un1}) (\ref{eqn:un2}).
$\hfill$

During the deployment, we solve a Control Barrier Function-Quadratic Program (CBF-QP) using \Cref{eq:cbf optimization problem} that minimizes the deviation from the desired surgical policy control $a_{des}$.
The final safe action is constructed by replacing $a_{des}$ with the optimized action.
The detailed algorithm of the whole framework can be found in \Cref{alg:ssp}.


\section{Implementation}
In this section, we present the detailed implementation of the CLF and CBF according to different surgical applications.

\subsection{Reference Path Following with CLF}
The reference Path following is implemented as follows. Given \Cref{eq:clf optimization problem}, we solve a Quadratic Program (QP) at each time step, which is further converted into a standard solver form: 
\begin{equation}
    \min \frac{1}{2}a^T P a + q^T a \text{ s.t. } Ga \le h
\end{equation}

The objective $\|a\|^2$ is represented by setting the cost matrix $P=\mathbf{I}$ (the identity matrix) and the cost vector $q = \mathbf{0}$.
The convergence constraint $L_f V(s) + L_g V(s) a + \beta V(s) \leq 0$ is rearranged into the linear inequality $Ga \leq h$, where $G = L_g V(s)$ and $h = -L_f V(s) - \beta V(s)$.
We use the Lyapunov function defined in \Cref{eq:lyapunov_function}.
The primary hyperparameter for this optimization problem is the gain $\beta$.

\subsection{Behavior CBF and Spatial CBF}
\begin{figure*}[t]
    \centering
    \begin{tabular}{@{} c c c @{}}
        
        (a) & 
        \raisebox{-0.5\height}{\includegraphics[width=0.42\linewidth]{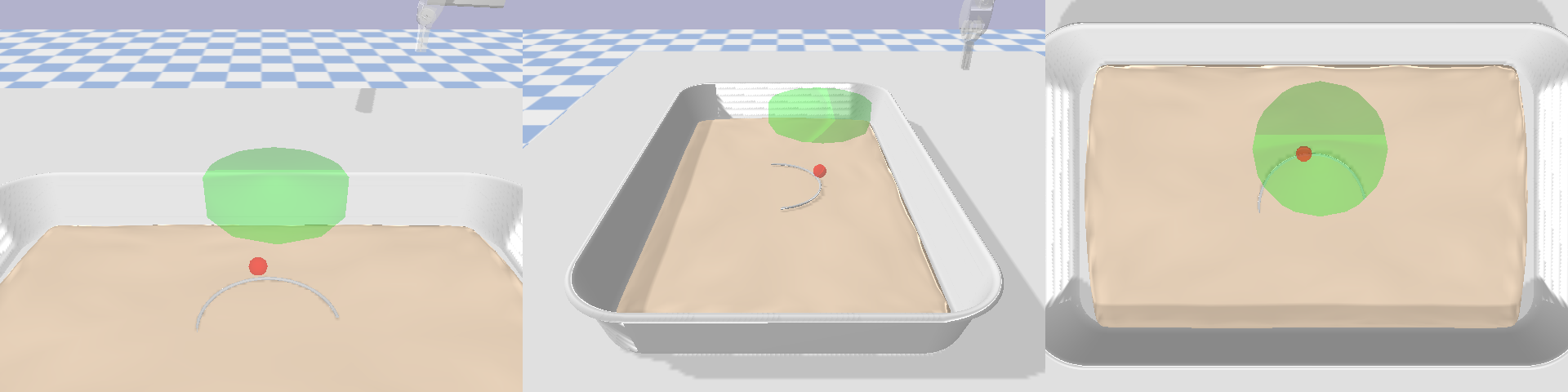}} & 
        \raisebox{-0.5\height}{\includegraphics[width=0.42\linewidth]{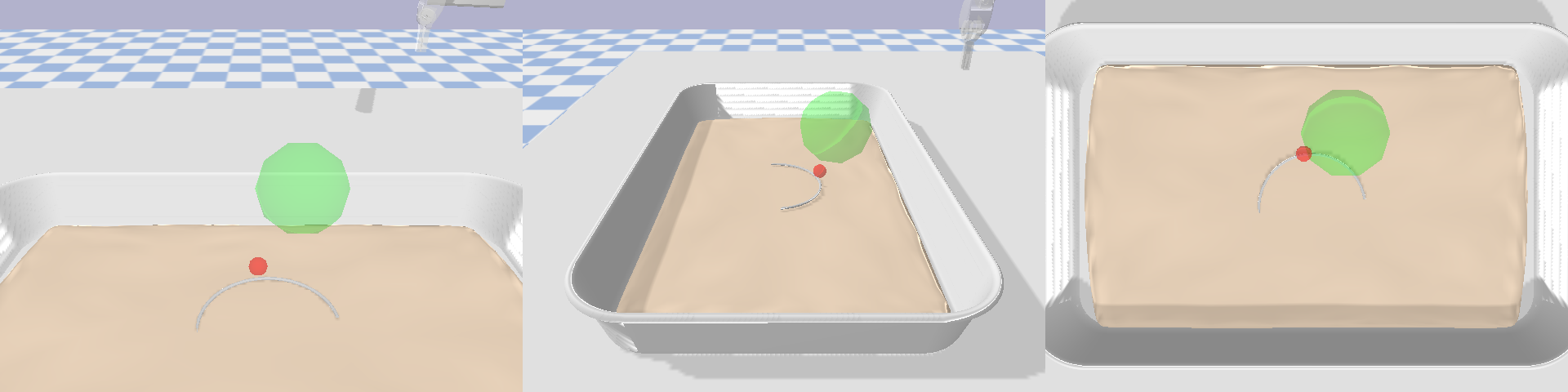}} \\
        \addlinespace 
        
        (b)  & 
        \raisebox{-0.5\height}{\includegraphics[width=0.42\linewidth]{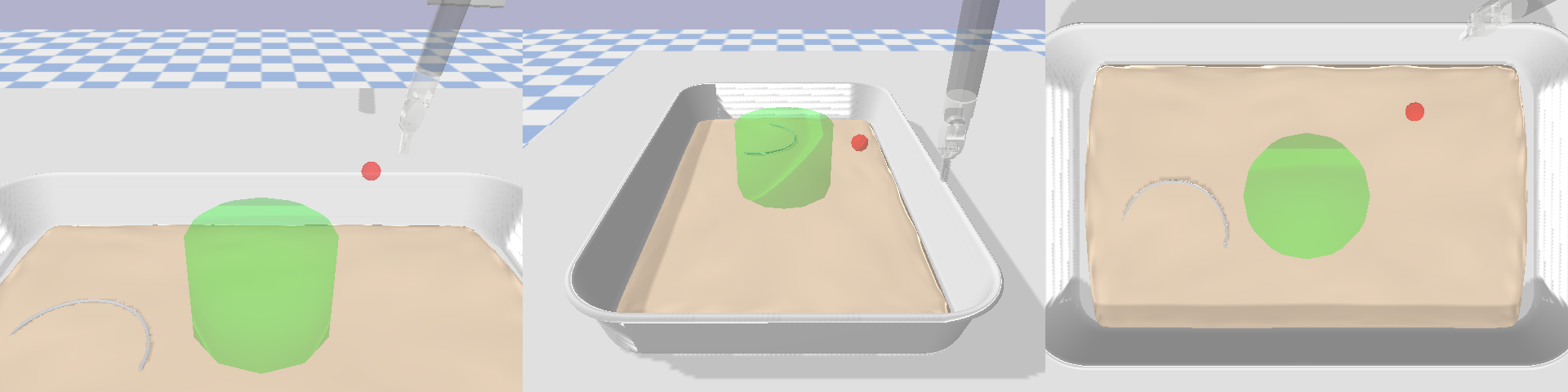}} & 
        \raisebox{-0.5\height}{\includegraphics[width=0.42\linewidth]{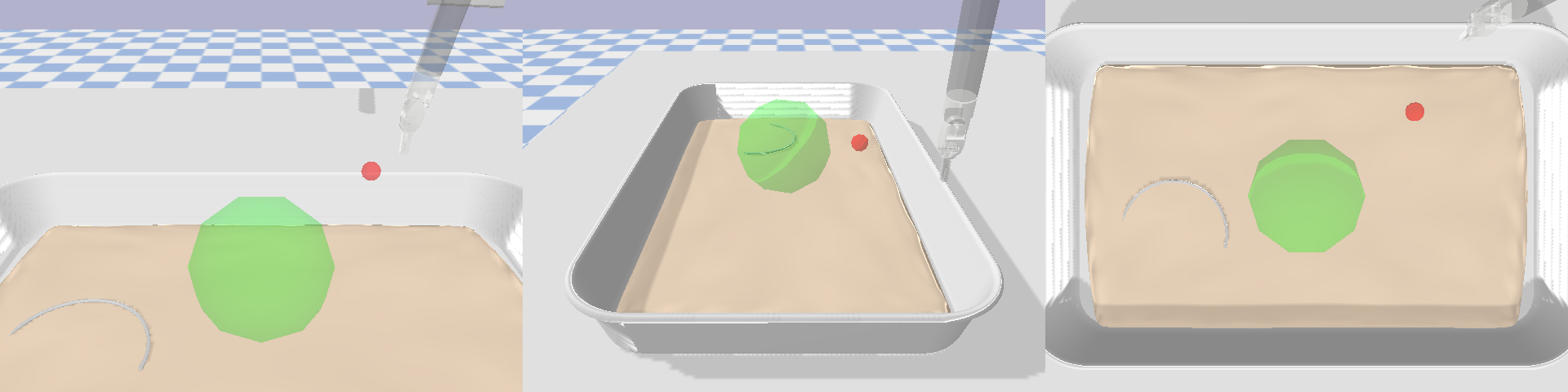}} \\
        \addlinespace
        
        (c) & 
        \raisebox{-0.5\height}{\includegraphics[width=0.42\linewidth]{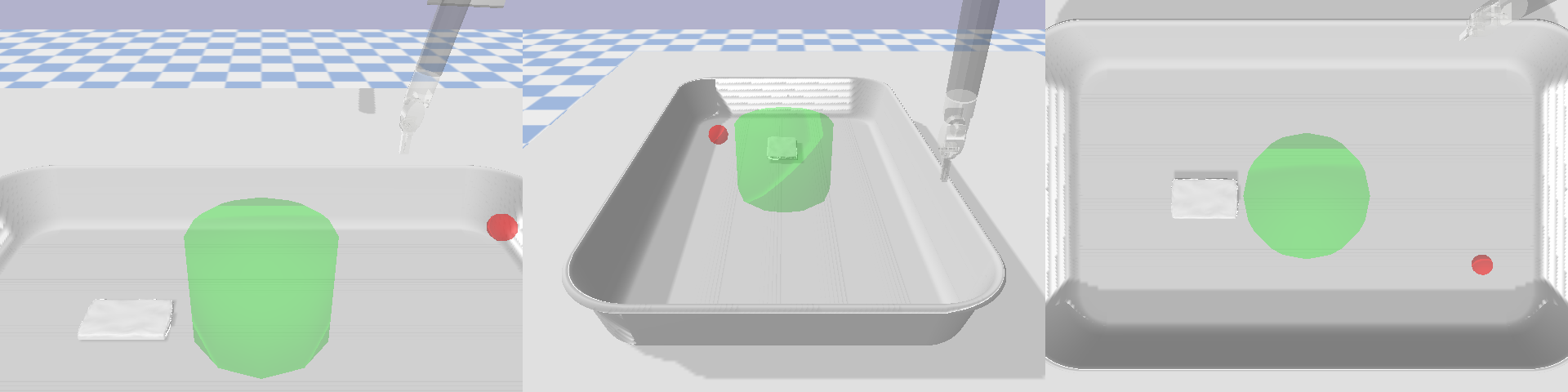}} & 
        \raisebox{-0.5\height}{\includegraphics[width=0.42\linewidth]{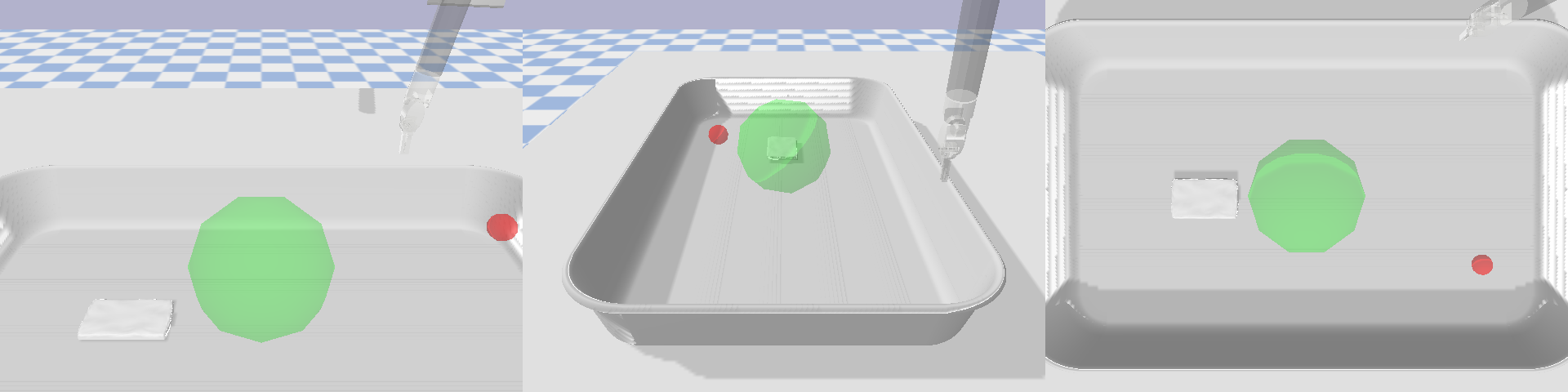}} \\
        \addlinespace
        
        (d) & 
        \raisebox{-0.5\height}{\includegraphics[width=0.42\linewidth]{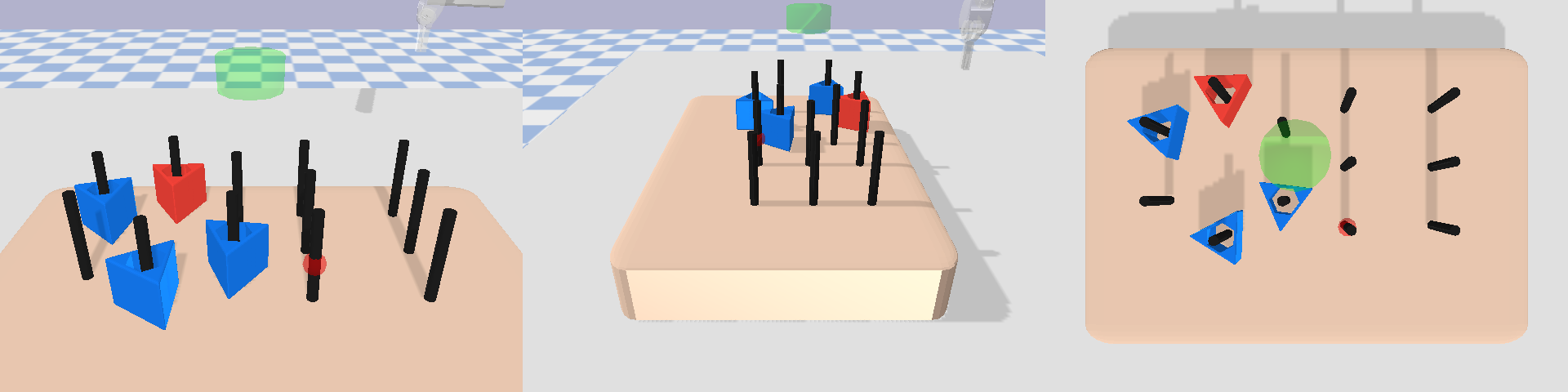}} & 
        \raisebox{-0.5\height}{\includegraphics[width=0.42\linewidth]{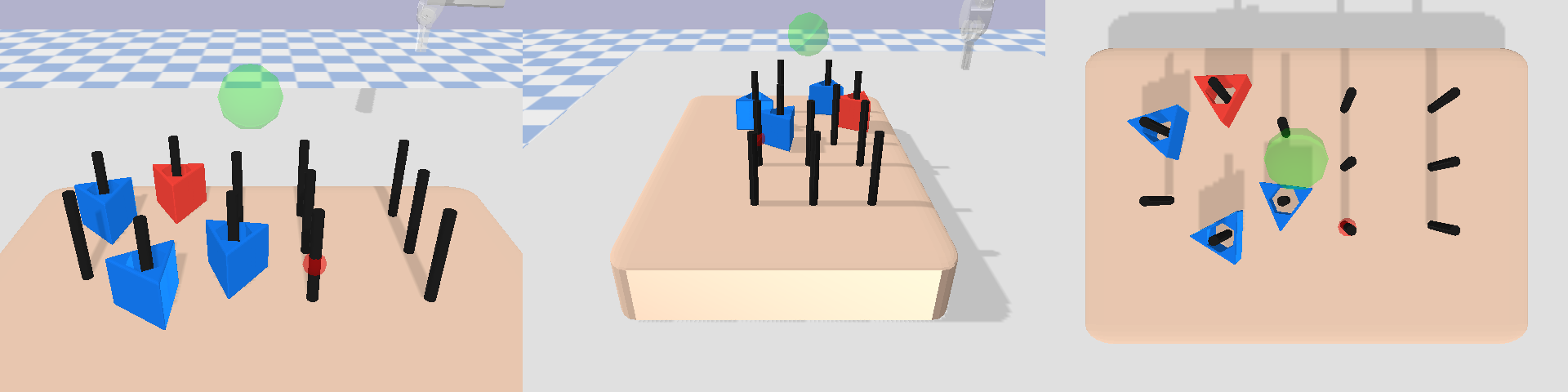}} \\
        \addlinespace
        
        & \textbf{Cylinder No-Go Zones} & \textbf{Sphere No-Go Zones} \\
        
    \end{tabular}
    
    \vspace{0.5em} 
    
    \caption{\textbf{Surgical Environments with Constraints} in SurRoL used to evaluate safety. The rows correspond to the four tasks: (a) NeedleReach, (b) NeedlePick, (c) GauzeRetrieve, and (d) PegTransfer. The columns illustrate the two different No-Go Zones geometries (Cylinder vs. Sphere).}
    \label{fig:constrained_tasks}
\end{figure*}

As mentioned in \Cref{method:cbf}, we consider two types of constraints:
behavioral constraint and spatial constraint.
The behavioral constraint is defined in \Cref{eq:task space} and the corresponding barrier function is defined in \Cref{eq:task space barrier function}.
Here, we set $d=0.5$ in our experiments.
For spatial constraint, we introduce a static geometric no-go zone (e.g., a sphere or cylinder) into the workspace, as shown in \Cref{fig:constrained_tasks}.
Here, we only consider constraints on the position of the robot.
The robust CBF safety controller is implemented by solving the QP defined in \Cref{eq:cbf optimization problem} at each time step.
This can also be converted into the standard solver form 
\begin{equation}
    \min \frac{1}{2}u^T P u + q^T u \text{ s.t. } Gu \le h
\end{equation}

The objective $\|u - u_{des}\|^2$ is represented by setting the cost matrix $P=\mathbf{I}$ (the identity matrix) and the cost vector $q = -u_{des}$.
The safety constraint $L_fb(y)+L_gb(y)u - \left|\frac{db(y)}{dy}\right| E_{\dot s}+\gamma b(y) \geq 0$ is rearranged into the linear inequality $Gu \leq h$, where $G = -L_g b(y)$ and $h = L_f b(y) - \left|\frac{db(y)}{dy}\right|E_{\dot s}+ \gamma b(y)$.
Similarly, we rearrange the behavioral constraint into a standard QP form.
The primary hyperparameter for this optimization problem is the gain $\gamma$.

To construct the geometry-specific no-go zones, we formally define each CBF as a continuously differentiable function $b(x)$ such that the safe set $\mathcal{C}$ is characterized by $b(x) \geq 0$.

\noindent \textbf{Sphere No-Go Zone}
For a sphere no-go zone with center $x_c \in \mathbb{R}^3$ and radius $r$, the safe region is the space outside the sphere. The barrier function is therefore the squared distance to the center minus the squared radius. For an end-effector at position $x$: $$b(x) = \|x-x_{c}\|^2 -r^2$$

\noindent \textbf{Cylinder No-Go Zone}
For a cylinder, safety is defined by a composite constraint: the end-effector is considered safe if it lies outside the cylinder, which occurs when either (1) it is radially outside the curverd side surface, or (2) it is vertically above the top plane or below the bottom plane.
This logical "OR" condition is implemented by combining two separate barrier functions.
Let the cylinder be parameterized by a point $c \in \mathbb{R}^3$ on its central axis, a unit vector $v \in \mathbb{R}^3$ indicating its axis direction, its radius $r_{cyl}>0$ and its length $l_{cyl}>0$.

Therefore, we define a two-component barrier function: a radial barrier and a vertical barrier.
The radial barrier measures how far the point $x$ is from the lateral surface in the direction perpendicular to the axis.
The perpendicular distance from the point $x$ to the axis is the norm of the cross product between the vector $(x-c)$ and the axis vector $v$.
The barrier function is this distance minus the cylinder radius:
$$b_{\text{radial}}(x) = \|(x - c) \times v\| - r_{cyl}.$$
The value is non-negative when $x$ is outside the side surface of the cylinder.

The vertical barrier measures the vertical offset of $x$ relative to the center $c$.
The signed distance along the axis is is computed using the dot product $|(x - c) \cdot v|$.
So, to correctly represent safety outside the top and bottom caps, we must compare this axial distance to half the height:
$$b_{\text{vertical}}(x) = |(x - c) \cdot v|-\frac{l_{cyl}}{2}.$$
    
The final, composite barrier function is the maximum of these two components. The system is safe as long as $b(x) \ge 0$, which is true if either $b_{\text{radial}}$ or $b_{\text{vertical}}$ is non-negative.
$$b(x) = soft\max\left(b_{\text{radial}}(x), ~ b_{\text{vertical}}(x)\right)$$

\section{Experiments and Results}
\label{sec:exp}
\subsection{Surgical Tasks and Environments}
\begin{figure}[tb]
    \centering
    \includegraphics[width=0.24\linewidth]{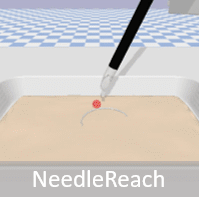}
    \includegraphics[width=0.24\linewidth]{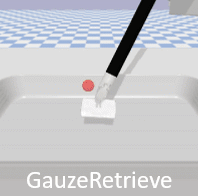}
    \includegraphics[width=0.24\linewidth]{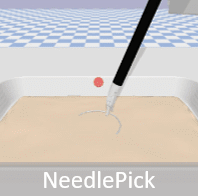}
    \includegraphics[width=0.24\linewidth]{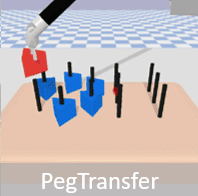}
    \caption{\textbf{Surgical Environments:} The four unconstrained simulation environments in SurRoL used for evaluation.}
    \label{fig:surrol}
\end{figure}

We validate our proposed SSP framework within the SurRoL simulation platform \citep{xu2021surrol} and real-world dVRK system.
SurRoL is an open-source, dVRK-compatible environment designed for surgical robot learning.
We evaluate our method on three typical reference path following task (straight, circular and triangular lines) and four representative surgical manipulation tasks as shown in \Cref{fig:surrol}:\\
- \textbf{NeedleReach:} The goal is to move the
    jaw tip to the location slightly above a needle.\\
-  \textbf{NeedlePick:} The robot is required to move to a needle, grasp it, and transport it to a goal position (marked by a red sphere).\\
- \textbf{GauzeRetrieve:} The robot is tasked with moving towards a piece of gauze, grasping it, and retrieving it to a goal position.\\
- \textbf{PegTransfer:} The robot needs to move to a block (marked as red) from one peg, grasp it, and move it to another peg (marked by a red sphere).

In these environments, the state $s$ is the 6D pose (position and orientation) of the end-effector and the action space contains 3D position control and yaw angle control.
For each task, a dataset of expert demonstrations $\mathcal{D}=\{\tau_i, i=1,...,N\}$ are provided, where each trajectory $\tau_i$ consists of a sequence of states and actions $\tau_i = (s_0^i, a^i_0,...,s_H^i)$.

\subsection{Model Parameters}
We introduce the implementation details for Neural ODE dynamics model learning. In the control-affine form $\dot{s} = f_{\eta}(s)+g_{\eta}(s)a$, we parameterize both the drift $f_\eta(s) \in \mathbb{R}^4$ and the control matrix $g_\eta(s) \in \mathbb{R}^{6*4}$ using a single Multi-Layer Perceptron (MLP) \citep{haykin1994neural} with parameters $\eta$.
The MLP consists of an input layer of 4 neurons (for $s \in \mathbb{R}^4$), one hidden layer of 64 dimensions with a GELU \citep{Hendrycks2016GaussianEL} activation, and a linear output layer of 30 neurons.
This output of 30 dimensions is then decomposed into $f_\eta(s)$ (the first 6 elements) and $g_\eta(s)$ (the remaining 24 elements). The Neural ODE model is trained by sampling a batch of $B=20$ trajectory segments, each of length $h=10$ timesteps, from the 100 available trajectories.
Given an initial state $s_0$, we performed a multi-step prediction rollout, where the predicted state $\hat{s}_t$ is used as the initial condition for predicting $\hat{s}_{t+1}$ using \Cref{eq:neural ode integral}.
%
This integral is computed using a differential equation solver, with a discretization time of 0.1s.
The network parameters $\eta$ are optimized by minimizing the Mean Absolute Error (L1 Loss) between the predicted trajectory $\{\hat{s}_i\}$ and the ground-truth trajectory $\{s_i\}$:
$$\mathcal{L}(\eta) = \frac{1}{B \cdot h} \sum_{j=1}^B \sum_{t=1}^h \| \hat{s}_t^{(j)} - s_t^{(j)} \|_1$$

This multi-step rollout loss is crucial for ensuring the long-term stability of the learned dynamics. The network is trained for 200 epochs using the RMSprop optimizer with a learning rate of $1 \times 10^{-3}$.
\begin{figure}[t]
    \centering
    \subfloat[Trajectory\label{fig:traj_circle}]{
        \includegraphics[width=0.48\linewidth]{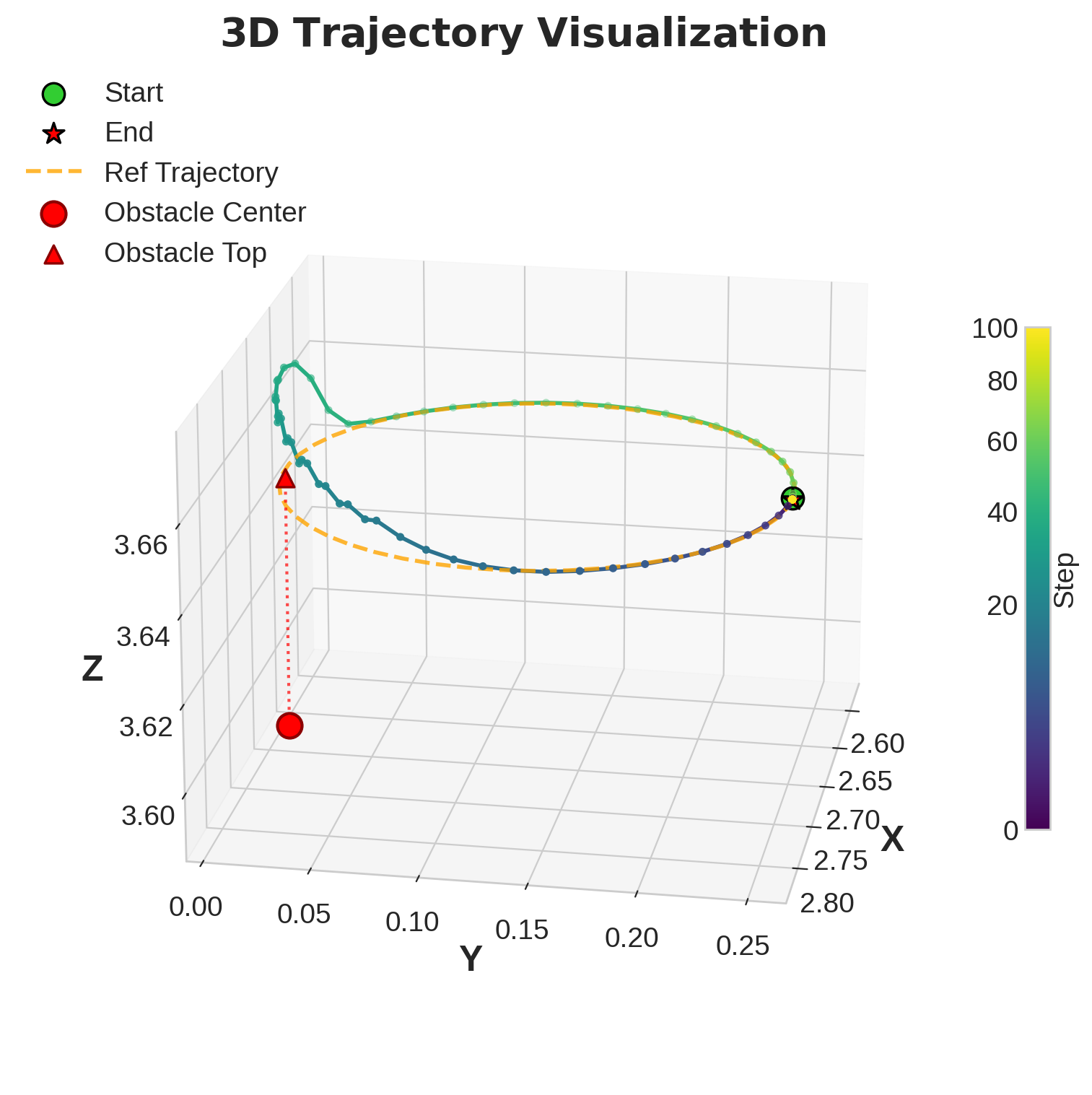}
    }
    \subfloat[Trajectory \label{fig:traj_cylinder}]{
        \includegraphics[width=0.48\linewidth]{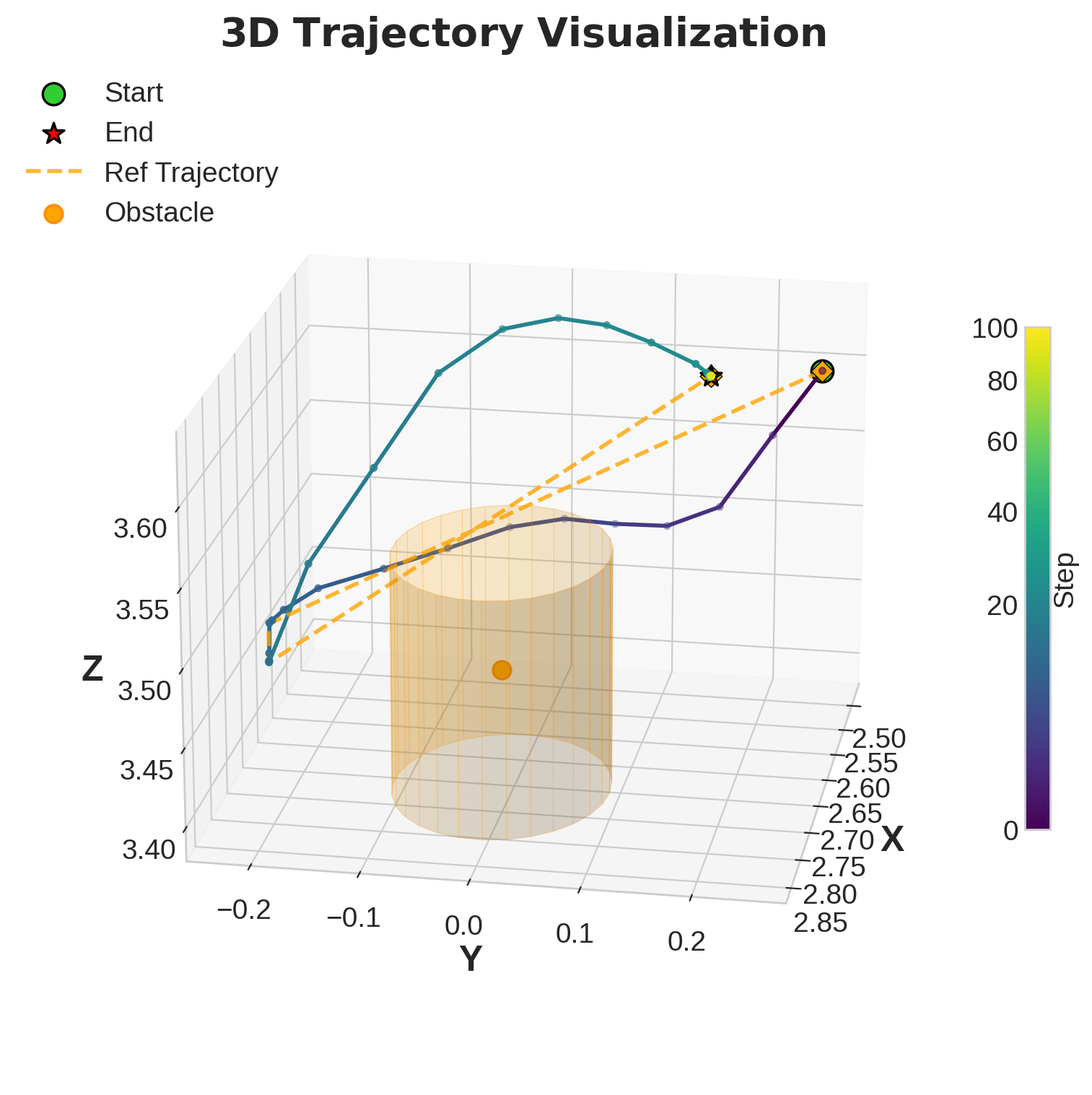} 
    }
    \hfill 
    \subfloat[Safe Margin\label{fig:cbf_circle}]{
        \includegraphics[width=0.48\linewidth]{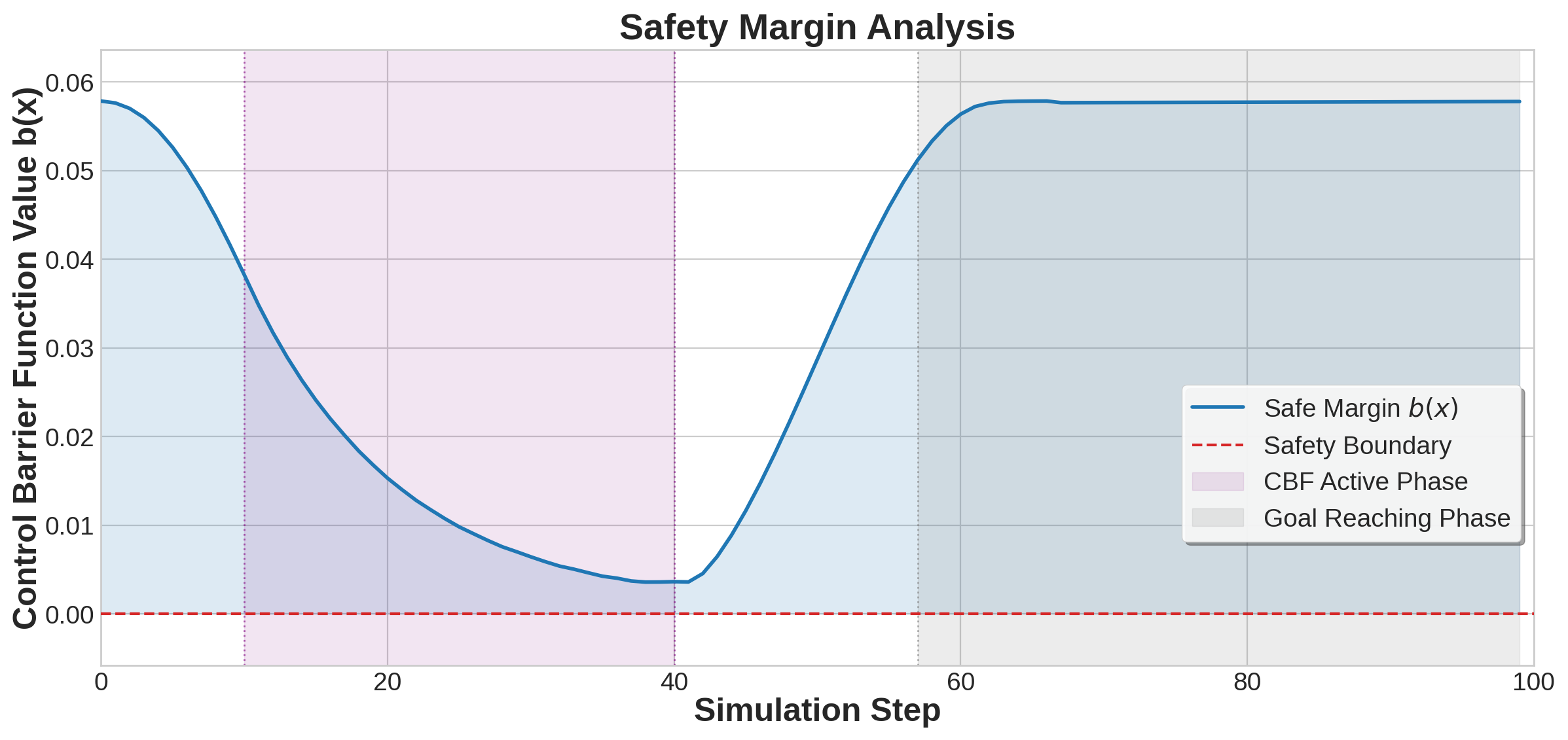}
    }
    \subfloat[Safe Margin\label{fig:cbf_cylinder}]{
        \includegraphics[width=0.48\linewidth]{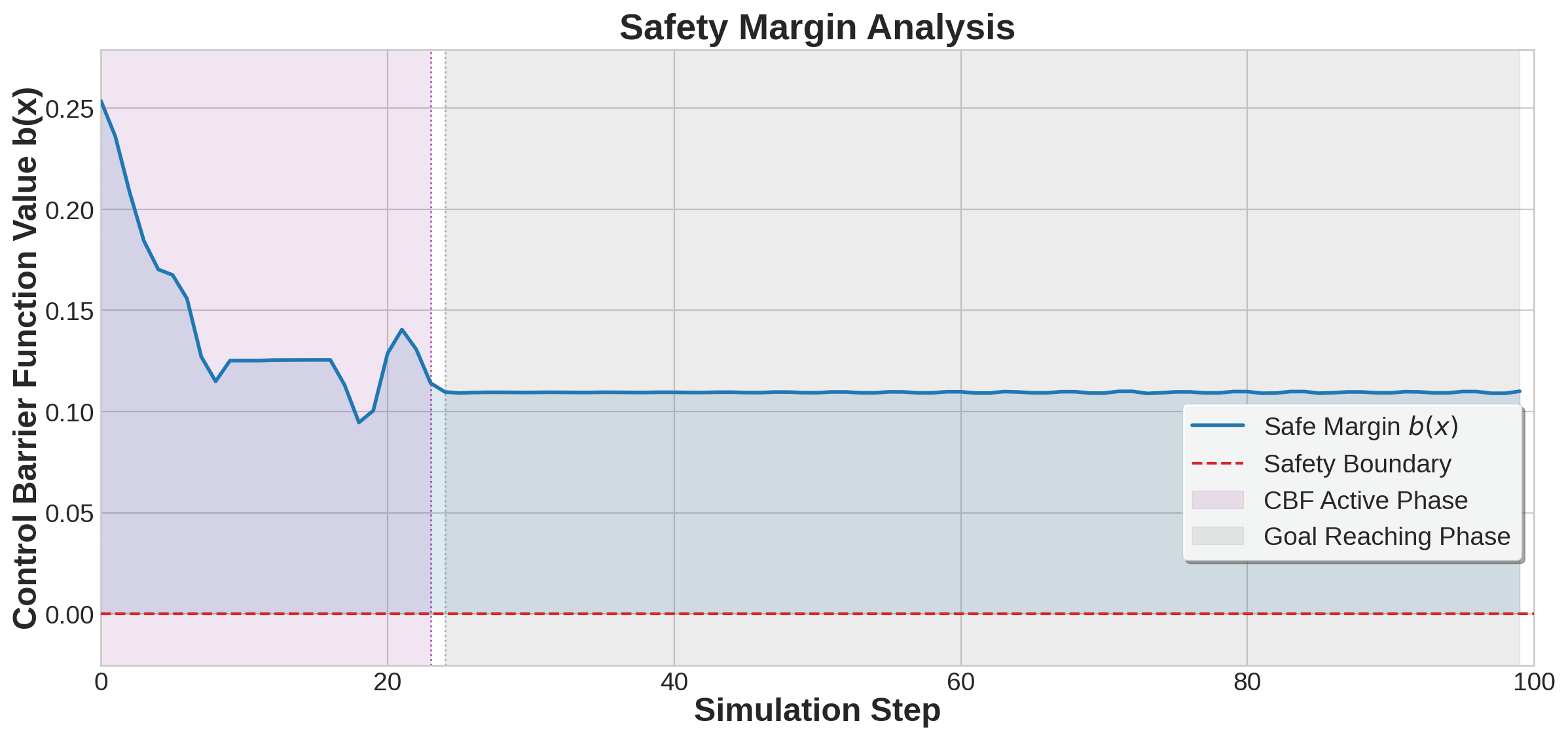}
    }
    
    \caption{\textbf{Visualization of the Trajectory and Corresponding Safe Margin:} 
    (a) The circular path following task and (c) the safe margin along the path.
    With CLF combined with CBF, the agent successfully follow the circular path while avoid the red sphere no-go zone.
    (b) Path following for NeedlePick task and (d) the safe margin along the path.
    With CLF combined with CBF, the agent successfully finish the \textit{NeedlePick} task while avoid the cylinder no-go zone. 
    The margin values remain strictly positive ($b(x) > 0$), quantitatively verifying that the safety constraints are strictly satisfied throughout the execution.}
    \label{fig:trajectories_visualization}
\end{figure}

\subsection{Experiment with Toy Examples}
We first evaluate the path following performance achieved by CLF using three representative paths: straight, circular, and triangular paths.
For each path, the mean deviation between the executed path and the reference path are recorded in \Cref{tab:combined_tracking_results}.

We further assess our framework, instantiated with a CLF policy, on facing an sphere no-go zone along the tracking path.
In this configuration, the evaluation metrics adapts based on whether the safety filter is activated.
Specifically, when the safety filter is inactive, which means the nominal action is not modified by the CBF optimization, we measure the deviation with respect to the reference path.
Conversely, during active CBF intervention, the deviation is measured relative to the no-go zone boundary.
Deviations corresponding to different phases are reported in \Cref{tab:combined_tracking_results}.

Meanwhile, we visualize the path of using CLF to track a circular line and using CBF to avoid a sphere no-go zone, as shown in \Cref{fig:trajectories_visualization}.
The robot can track the reference path well while deviate from it to successfully avoid the no-go zone.
The safe margins along this path are also recorded in \Cref{fig:trajectories_visualization} to show the change of it during different phases in this task.

\begin{table}[t]
    \centering
    \caption{\textbf{Comprehensive Reference Path Following Results:} Average trajectory tracking deviation from the reference path under CLF control and CLF CBF control.}
    \label{tab:combined_tracking_results}
    \small 
    \setlength{\tabcolsep}{3pt} 
    \resizebox{0.9\linewidth}{!}{
    \begin{tabular}{@{} c c c c c c @{}}
        \toprule
        & & \shortstack[c]{\textbf{CLF Only}} & \multicolumn{3}{c}{\textbf{Combined CLF-OKCBF Control}} \\
        \cmidrule(lr){3-3} \cmidrule(l){4-6}
        \shortstack[l]{\textbf{Path} \\ \textbf{Type}} & \shortstack[c]{\textbf{Len.} \\ \textbf{(m)}} & \shortstack[c]{\textbf{Avg. Dev.} \\ \textbf{($10^{-4}$ m)}} & \shortstack[c]{\textbf{CLF Dev.} \\ \textbf{($10^{-4}$ m)}} & \shortstack[c]{\textbf{CBF Dev.} \\ \textbf{($10^{-2}$ m)}} & \shortstack[c]{\textbf{Total Dev.} \\ \textbf{($10^{-3}$ m)}} \\
        \midrule
        Straight   & $0.35$ & $0.939$ & $1.00$ & $7.89$  & $3.23$  \\
        Circular   & $0.75$ & $4.24$  & $3.00$ & $0.230$ & $0.341$ \\
        Triangular & $0.30$ & $4.34$  & $4.00$ & $3.05$  & $1.35$  \\
        \bottomrule
    \end{tabular}
    }
\end{table}



Next, we investigate the sensitivity of the proposed framework to the hyperparameters governing CLF and CBF, specifically characterizing the trade-off between trajectory tracking deviation and safety enforcement.
We choose the straight line as the reference path for testing the hyperparameters in CLF and CLF with CBF.

First, we analyze the influence of the CLF gain parameter ($\beta$) on tracking precision.
As detailed in \Cref{tab:ablation_clf_cbf}, increasing $\beta$ initially improves performance, with the average tracking deviation reaching a minimum of $9.39 \times 10^{-5}$~m at $\beta = 15$.
However, further increasing the gain to $\beta = 20$ or $25$ results in a marginal increase in deviation.
This inflection point suggests that while higher gains theoretically accelerate convergence, excessive values amplify the epistemic uncertainty inherent in the learned Neural ODE dynamics.
Furthermore, high gains can introduce artifacts near the equilibrium due to discretization, leading to over-correction that degrades precision.

\begin{table}[t]
    \centering
    \caption{\textbf{Ablation Studies:} Impact of CLF Gain $\beta$ (Left) and CBF Gain $\gamma$ (Right) on System Performance.}
    \label{tab:ablation_clf_cbf}
    \begin{tabular}{c c @{\hspace{0.1cm}} c c} 
        \hline

        \multicolumn{2}{c}{\textbf{Ablation Study (CLF)}} & \multicolumn{2}{c}{\textbf{Ablation Study (CBF)}} \\
\cmidrule(lr){1-2} \cmidrule(lr){3-4}
\textbf{Beta ($\beta$)} & \shortstack[c]{\textbf{Avg. Tracking} \\ \textbf{Dev. (m)}} & \textbf{Gamma ($\gamma$)} & \shortstack[c]{\textbf{Avg. Safe} \\ \textbf{Margin (m)}} \\
        \hline
        $5$  & $2.75 \times 10^{-4}$ & $5$  & $9.94 \times 10^{-2}$ \\
        $10$ & $1.36 \times 10^{-4}$ & $\mathbf{10}$ & $\mathbf{7.89 \times 10^{-2}}$ \\
        $\mathbf{15}$ & $\mathbf{9.39 \times 10^{-5}}$ & $15$ & $7.35 \times 10^{-2}$ \\
        $20$ & $1.26 \times 10^{-4}$ & $20$ & $5.85 \times 10^{-2}$ \\
        $25$ & $1.34 \times 10^{-4}$ & $25$ & $4.83 \times 10^{-2}$ \\
        \hline
    \end{tabular}
\end{table}

Second, we perform an ablation study on the CBF gain parameter ($\gamma$) to evaluate its impact on task deviation and collision avoidance (\Cref{tab:ablation_clf_cbf}).
The results demonstrate a fundamental trade-off between the conservativeness of the safety filter and the robustness of the safety guarantee.
Lower values of $\gamma$ (e.g., $\gamma=5$) strictly enforce forward invariance of the safe set and induces a higher average safe margin of $9.94 \times 10^{-2}$~m due to the early, more restrictive activation of the barrier constraint.
However, too small $\gamma$ values are found to jeopardize task performance in dynamic scenarios.
As $\gamma$ increases, the safe margin consistently decreases, reaching $4.83 \times 10^{-2}$~m at $\gamma=25$.
Correspondingly, the larger gain permits the system to approach the safety boundary too aggressively, leaving insufficient control authority to compensate for system latency or model mismatch, which may break the safety constraints.

\subsection{Experiment in SurRoL Simulation Environment}

\noindent \textbf{Experimental Setups} We collect a dataset of 100 demonstrations per task from the SurRoL simulation platform, where each trajectory contains 100 steps.
We evaluate the effectiveness of our proposed framework in achieving safe and successful task execution.
In safety-critical systems, an execution is only considered successful if the task goal is achieved without violating any prescribed safety constraint.
We implement our framework (SSP) with three surgical policies: demonstration-guided RL policy (DEX\citep{dex}), diffusion policy \citep{dsp} and CLF-based path follower.

\begin{table*}[p]
\caption{\textbf{Evaluation Results.} Mean success rate (with / without violation), collision rate, inference time, and safety margin. Results are averaged over 3 seeds with 20 rollouts each. $\uparrow$ / $\downarrow$ indicate higher / lower is better.
The quantitative results demonstrate the critical efficacy of the proposed framework in enforcing safety without severely compromising task performance.}
\label{tab:success_rate}
\centering
\setlength{\tabcolsep}{4pt}
\renewcommand{\arraystretch}{1.15}
\resizebox{\textwidth}{!}{
\begin{tabular}{lccccccccc}
\toprule
\multirow{3}{*}{\textbf{Constraint}} & \multirow{3}{*}{\textbf{Task}} & \multirow{3}{*}{\textbf{Method}} & \multicolumn{3}{c}{\textbf{Task Performance}}  & \multicolumn{4}{c}{\textbf{Safety / Efficiency}} \\
\cmidrule(lr){4-6} \cmidrule(lr){7-10}
& & & \multicolumn{2}{c}{\textbf{Mean Success Rate$\uparrow$}} & \multirow{2}{*}{\textbf{Collision}$\downarrow$} &\textbf{Time$\downarrow$} &\textbf{Safe Margin $\uparrow$} & \textbf{$\dot s$ Error $\downarrow$} & \textbf{$s$ Error $\downarrow$}\\
\cmidrule(lr){4-5} 
& & & \textbf{(w/ Violation)} & \textbf{(w/o Violation)} & &\textbf{(ms)} & \textbf{($\times 10^{-3}$)} & \textbf{($\times 10^{-3}$ )}&\textbf{($\times 10^{-4}$ )}\\
\midrule
\multirow{24}{*}{\shortstack[c]{\textbf{Cylinder} \\ \textbf{No-Go} \\ \textbf{Zones}}} & \multirow{6}{*}{NeedleReach} 
& CLF \cite{ames2012control} & $1.00\!\pm\!0.00$ & $0.08\!\pm\!0.02$ & $0.92\!\pm\!0.02$ & $3.33\!\pm\!0.28$ & $-15.5\!\pm\!1.47$ & $1299\!\pm\!7.94$&$1295\!\pm\!7.87$\\
& & DP \cite{dsp} & $1.00\!\pm\!0.01$ & $0.42\!\pm\!0.03$ & $0.58\!\pm\!0.03$ & $6.96\!\pm\!0.36$ & $-1.20\!\pm\!1.61$ & $\mathbf{4.83\!\pm\!0.18}$&$\mathbf{4.79\!\pm\!0.17}$\\
& & DEX \cite{dex} & $1.00\!\pm\!0.00$ & $0.50\!\pm\!0.00$ & $0.50\!\pm\!0.00$ & $\mathbf{1.73\!\pm\!0.09}$ & $1.71\!\pm\!0.89$ & $6.60\!\pm\!0.29$&$6.68\!\pm\!0.33$\\
& & \textbf{SSP-CLF (Proposed)} & $1.00\!\pm\!0.00$ & $0.27\!\pm\!0.09$ & $0.17\!\pm\!0.02$ & $5.17\!\pm\!0.09$ & $2.24\!\pm\!0.79$ & $1074\!\pm\!5.80$&$1071\!\pm\!5.93$\\
& & \textbf{SSP-DP (Proposed)} & $1.00\!\pm\!0.01$ & $0.50\!\pm\!0.05$ & $0.12\!\pm\!0.08$ & $9.37\!\pm\!0.03$ & $6.33\!\pm\!1.86$ &$9.73\!\pm\!0.44$&$9.75\!\pm\!0.44$\\
& & \textbf{SSP-DEX (Proposed)} & $1.00\!\pm\!0.00$& $\mathbf{1.00\!\pm\!0.00}$ & $\mathbf{0.00\!\pm\!0.00}$ & $4.30\!\pm\!0.08$ & $\mathbf{44.0\!\pm\!1.88}$ & $26.3\!\pm\!0.61$&$26.9\!\pm\!0.56$\\
\cmidrule(lr){2-10}
& \multirow{6}{*}{GauzeRetrieve} 
& CLF \cite{ames2012control} & {$1.00\!\pm\!0.00$} & $0.02\!\pm\!0.02$ & $0.98\!\pm\!0.02$ & $3.30\!\pm\!0.29$ & $-31.9\!\pm\!1.69$ & $1642\!\pm\!18.0$&$1635\!\pm\!19.0$\\
& & DP \cite{dsp} & {$0.73\!\pm\!0.08$} & $0.13\!\pm\!0.03$ & $0.75\!\pm\!0.05$ & $6.73\!\pm\!0.07$ & $-7.40\!\pm\!2.00$ & $\mathbf{5.33\!\pm\!0.61}$&$\mathbf{5.37\!\pm\!0.61}$\\
& & DEX \cite{dex} & {$0.72\!\pm\!0.12$} & $0.12\!\pm\!0.06$ & $0.72\!\pm\!0.09$ & $\mathbf{1.63\!\pm\!0.09}$ & $-2.08\!\pm\!7.42$ & $9.09\!\pm\!4.67$&$9.16\!\pm\!4.65$\\
& & \textbf{SSP-CLF (Proposed)} & $1.00\!\pm\!0.00$ & $\mathbf{0.98\!\pm\!0.02}$ & $\mathbf{0.00\!\pm\!0.00}$& $6.97\!\pm\!0.05$ & $22.0\!\pm\!0.54$ & $1309\!\pm\!63.1$&$1303\!\pm\!62.6$\\
& & \textbf{SSP-DP (Proposed)} & $0.73\!\pm\!0.08$ & $0.20\!\pm\!0.00$ & $0.18\!\pm\!0.13$ & $9.33\!\pm\!0.06$ & $3.77\!\pm\!0.32$ & $34.6\!\pm\!8.46$&$34.6\!\pm\!8.46$\\
& & \textbf{SSP-DEX (Proposed)} & $0.72\!\pm\!0.12$ & $0.68\!\pm\!0.06$ & $\mathbf{0.00\!\pm\!0.00}$ & $3.57\!\pm\!0.05$ & $\mathbf{68.7\!\pm\!5.87}$ & $11.4\!\pm\!3.99$&$11.5\!\pm\!3.99$\\
\cmidrule(lr){2-10}
& \multirow{6}{*}{NeedlePick} 
& CLF \cite{ames2012control} & {$0.78\!\pm\!0.12$} & $0.00\!\pm\!0.00$ & $1.00\!\pm\!0.00$ & $4.30\!\pm\!0.16$ & $-54.7\!\pm\!6.63$ & $1602\!\pm\!12.5$&$1608\!\pm\!12.4$\\
& & DP \cite{dsp} & {$0.99\!\pm\!0.02$} & $0.00\!\pm\!0.00$ & $1.00\!\pm\!0.00$ & $6.86\!\pm\!0.04$ & $-27.9\!\pm\!2.00$ & $\mathbf{5.22\!\pm\!0.25}$&$\mathbf{5.14\!\pm\!0.23}$\\
& & DEX \cite{dex} & {$0.94\!\pm\!0.05$} & $0.00\!\pm\!0.00$ & $0.98\!\pm\!0.02$ & $\mathbf{2.30\!\pm\!0.42}$ & $-40.0\!\pm\!2.40$ & $8.96\!\pm\!3.09$&$8.87\!\pm\!3.11$\\
& & \textbf{SSP-CLF (Proposed)} & $0.78\!\pm\!0.12$ & $\mathbf{0.88\!\pm\!0.02}$ & $\mathbf{0.00\!\pm\!0.00}$ & $7.47\!\pm\!0.13$ & $11.4\!\pm\!0.50$ & $1015\!\pm\!12.5$&$1019\!\pm\!12.6$\\
& & \textbf{SSP-DP (Proposed)} & $0.99\!\pm\!0.02$ & $0.12\!\pm\!0.03$ & $0.05\!\pm\!0.00$ & $10.2\!\pm\!0.10$ & $4.17\!\pm\!0.11$ & $23.7\!\pm\!7.92$&$23.6\!\pm\!7.96$\\
& & \textbf{SSP-DEX (Proposed)} & $0.94\!\pm\!0.05$ & $0.87\!\pm\!0.02$ & $\mathbf{0.00\!\pm\!0.00}$ & $5.03\!\pm\!0.62$ & $\mathbf{27.8\!\pm\!1.48}$ & $11.5\!\pm\!2.14$&$11.6\!\pm\!2.14$\\
\cmidrule(lr){2-10}
& \multirow{6}{*}{PegTransfer} 
& CLF \cite{ames2012control} & {$0.35\!\pm\!0.04$} & $0.12\!\pm\!0.06$ & $0.68\!\pm\!0.05$ & $3.60\!\pm\!0.25$ & $-3.08\!\pm\!1.51$ & $1155\!\pm\!35.2$&$1154\!\pm\!34.9$\\
& & DP \cite{dsp} & {$0.99\!\pm\!0.02$} & $0.00\!\pm\!0.00$ & $1.00\!\pm\!0.00$ & $6.80\!\pm\!0.07$ & $-18.9\!\pm\!0.06$ & $\mathbf{6.01\!\pm\!0.82}$&$\mathbf{6.01\!\pm\!0.82}$\\
& & DEX \cite{dex} & {$0.73\!\pm\!0.20$} & $0.32\!\pm\!0.13$ & $0.57\!\pm\!0.17$ & $\mathbf{1.70\!\pm\!0.00}$ & $1.23\!\pm\!4.11$ & $31.3\!\pm\!8.19$&$31.3\!\pm\!8.19$\\
& & \textbf{SSP-CLF (Proposed)} & $0.35\!\pm\!0.04$ & $0.20\!\pm\!0.04$ & $0.08\!\pm\!0.02$ & $6.50\!\pm\!0.65$ & $17.5\!\pm\!1.26$ & $949\!\pm\!26.2$&$948\!\pm\!26.6$\\
& & \textbf{SSP-DP (Proposed)} & $0.99\!\pm\!0.02$ & $\mathbf{0.65\!\pm\!0.09}$ & $0.10\!\pm\!0.05$ & $9.50\!\pm\!0.03$ & $1.37\!\pm\!0.76$ & $50.8\!\pm\!20.2$&$50.8\!\pm\!20.2$\\
& & \textbf{SSP-DEX (Proposed)} & $0.73\!\pm\!0.20$ & $\mathbf{0.65\!\pm\!0.08}$ & $\mathbf{0.00\!\pm\!0.00}$ & $4.20\!\pm\!0.08$ & $\mathbf{51.4\!\pm\!2.59}$ & $26.4\!\pm\!3.45$&$26.4\!\pm\!3.44$\\
\midrule
\multirow{24}{*}{\shortstack[c]{\textbf{Sphere} \\ \textbf{No-Go} \\ \textbf{Zones}}} & \multirow{6}{*}{NeedleReach} 
& CLF \cite{ames2012control} & {$1.00\!\pm\!0.00$} & $0.08\!\pm\!0.02$ & $0.92\!\pm\!0.02$ & $3.20\!\pm\!0.22$ & $-2.82\!\pm\!0.36$ & $1299\!\pm\!7.94$&$1295\!\pm\!7.87$\\
& & DP \cite{dsp} & {$1.00\!\pm\!0.01$} & $0.18\!\pm\!0.08$ & $0.82\!\pm\!0.08$ & $6.47\!\pm\!0.05$ & $-1.63\!\pm\!0.50$ & $\mathbf{4.83\!\pm\!0.18}$&$\mathbf{4.79\!\pm\!0.17}$\\
& & DEX \cite{dex} & {$1.00\!\pm\!0.00$} & $0.30\!\pm\!0.04$ & $0.70\!\pm\!0.04$ & $\mathbf{1.63\!\pm\!0.09}$ & $-1.17\!\pm\!0.32$ & $6.60\!\pm\!0.29$&$6.68\!\pm\!0.33$\\
& & \textbf{SSP-CLF (Proposed)} & $1.00\!\pm\!0.00$ & $\mathbf{1.00\!\pm\!0.00}$ & $\mathbf{0.00\!\pm\!0.00}$ & $4.57\!\pm\!0.09$ & $4.23\!\pm\!0.37$ & $407\!\pm\!7.67$&$406\!\pm\!7.66$\\
& & \textbf{SSP-DP (Proposed)} & $1.00\!\pm\!0.01$ & $0.45\!\pm\!0.05$ & $0.07\!\pm\!0.03$ & $8.27\!\pm\!0.11$ & $0.80\!\pm\!0.36$ & $6.56\!\pm\!0.23$&$6.55\!\pm\!0.24$\\
& & \textbf{SSP-DEX (Proposed)} & $1.00\!\pm\!0.00$ & $\mathbf{1.00\!\pm\!0.00}$ & $\mathbf{0.00\!\pm\!0.00}$ & $3.47\!\pm\!0.05$ & $\mathbf{10.4\!\pm\!0.51}$ & $20.7\!\pm\!1.16$&$20.9\!\pm\!1.16$\\
\cmidrule(lr){2-10}
& \multirow{6}{*}{GauzeRetrieve} 
& CLF \cite{ames2012control} & {$1.00\!\pm\!0.00$} & $0.00\!\pm\!0.00$ & $1.00\!\pm\!0.00$ & $3.33\!\pm\!0.48$ & $-1.61\!\pm\!0.11$ & $1642\!\pm\!18.0$&$1635\!\pm\!19.0$\\
& & DP \cite{dsp} & {$0.73\!\pm\!0.08$} & $0.07\!\pm\!0.08$ & $0.83\!\pm\!0.18$ & $6.50\!\pm\!0.04$ & $-2.37\!\pm\!0.61$ &$5.33\!\pm\!0.61$&$5.37\!\pm\!0.23$\\
& & DEX \cite{dex} & {$0.74\!\pm\!0.12$} & $0.40\!\pm\!0.07$ & $0.50\!\pm\!0.04$ & $\mathbf{1.63\!\pm\!0.09}$ & $0.54\!\pm\!0.47$ & $8.97\!\pm\!4.51$&$9.05\!\pm\!4.49$\\
& & \textbf{SSP-CLF (Proposed)} & $1.00\!\pm\!0.00$ & $\mathbf{1.00\!\pm\!0.00}$ & $\mathbf{0.00\!\pm\!0.00}$ & $4.57\!\pm\!0.05$ & $13.1\!\pm\!0.47$ & $876\!\pm\!19.5$&$878\!\pm\!19.6$\\
& & \textbf{SSP-DP (Proposed)} & $0.73\!\pm\!0.08$ & $0.28\!\pm\!0.06$ & $0.02\!\pm\!0.03$ & $8.24\!\pm\!0.05$ & $1.07\!\pm\!0.06$ &$46.4\!\pm\!12.6$&$46.4\!\pm\!12.6$\\
& & \textbf{SSP-DEX (Proposed)} & $0.74\!\pm\!0.12$ & $0.82\!\pm\!0.02$ & $\mathbf{0.00\!\pm\!0.00}$ & $3.03\!\pm\!0.09$ & $\mathbf{24.2\!\pm\!0.06}$ & $\mathbf{4.73\!\pm\!0.13}$&$\mathbf{4.80\!\pm\!0.14}$\\
\cmidrule(lr){2-10}
& \multirow{6}{*}{NeedlePick} 
& CLF \cite{ames2012control} & {$0.78\!\pm\!0.12$} & $0.00\!\pm\!0.00$ & $1.00\!\pm\!0.00$ & $4.30\!\pm\!0.16$ & $-7.72\!\pm\!0.92$ & $1602\!\pm\!12.5$&$1608\!\pm\!12.4$\\
& & DP \cite{dsp} & {$0.99\!\pm\!0.02$} & $0.03\!\pm\!0.03$ & $0.97\!\pm\!0.03$ & $6.62\!\pm\!0.05$ & $-6.63\!\pm\!0.45$ & $\mathbf{5.22\!\pm\!0.26}$&$\mathbf{5.14\!\pm\!0.23}$\\
& & DEX \cite{dex} & {$0.94\!\pm\!0.05$} & $0.00\!\pm\!0.00$ & $0.98\!\pm\!0.02$ & $\mathbf{1.73\!\pm\!0.09}$ & $-7.91\!\pm\!0.41$ & $9.31\!\pm\!3.58$&$9.22\!\pm\!3.59$\\
& & \textbf{SSP-CLF (Proposed)} & $0.78\!\pm\!0.12$ & $\mathbf{0.87\!\pm\!0.09}$ & $\mathbf{0.00\!\pm\!0.00}$ & $6.73\!\pm\!0.13$ & $3.90\!\pm\!0.49$ & $554\!\pm\!12.3$&$556\!\pm\!12.3$\\
& & \textbf{SSP-DP (Proposed)} & $0.99\!\pm\!0.02$ & $0.12\!\pm\!0.03$ & $0.02\!\pm\!0.03$ & $9.29\!\pm\!0.09$ & $1.53\!\pm\!0.15$ &$8.34\!\pm\!1.90$&$8.25\!\pm\!1.89$\\
& & \textbf{SSP-DEX (Proposed)} & $0.94\!\pm\!0.05$ & $0.62\!\pm\!0.14$ & $\mathbf{0.00\!\pm\!0.00}$ & $3.70\!\pm\!0.14$ & $\mathbf{9.93\!\pm\!1.71}$ & $18.2\!\pm\!8.45$&$18.3\!\pm\!8.44$\\
\cmidrule(lr){2-10}
& \multirow{6}{*}{PegTransfer} 
& CLF \cite{ames2012control} & {$0.35\!\pm\!0.04$} & $0.17\!\pm\!0.06$ & $0.47\!\pm\!0.02$ & $3.53\!\pm\!0.29$ & $0.59\!\pm\!0.24$ & $1171\!\pm\!83.9$&$1169\!\pm\!83.4$\\
& & DP \cite{dsp} & {$0.99\!\pm\!0.02$} & $0.00\!\pm\!0.00$ & $1.00\!\pm\!0.00$ & $6.65\!\pm\!0.08$ & $-0.60\!\pm\!0.00$ &$\mathbf{5.92\!\pm\!0.55}$&$\mathbf{5.92\!\pm\!0.55}$\\
& & DEX \cite{dex} & {$0.73\!\pm\!0.20$} & $0.27\!\pm\!0.02$ & $0.58\!\pm\!0.02$ & $\mathbf{1.63\!\pm\!0.05}$ & $0.56\!\pm\!1.03$ & $29.4\!\pm\!3.31$&$29.4\!\pm\!3.31$\\
& & \textbf{SSP-CLF (Proposed)} & $0.35\!\pm\!0.04$ & $0.22\!\pm\!0.05$ & $\mathbf{0.00\!\pm\!0.00}$ & $5.87\!\pm\!0.62$ & $6.24\!\pm\!0.43$ & $480\!\pm\!39.7$&$480\!\pm\!39.7$\\
& & \textbf{SSP-DP (Proposed)} & $0.99\!\pm\!0.02$ & $\mathbf{0.97\!\pm\!0.06}$ & $\mathbf{0.00\!\pm\!0.00}$ & $8.38\!\pm\!0.15$ & $2.73\!\pm\!3.70$ &$65.7\!\pm\!5.73$&$65.7\!\pm\!5.73$\\
& & \textbf{SSP-DEX (Proposed)} & $0.73\!\pm\!0.20$ & $0.55\!\pm\!0.07$ & $\mathbf{0.00\!\pm\!0.00}$ & $3.47\!\pm\!0.05$ & $\mathbf{13.5\!\pm\!0.38}$ & $54.9\!\pm\!11.4$&$54.9\!\pm\!11.4$\\
\bottomrule
\end{tabular}
}
\end{table*}

\begin{table*}[htp]
\caption{\textbf{Performance of Different Learning Algorithms in Surgical Environments:} The quantitative results demonstrate the proposed framework achieves comparable performance compared to other unconstrained baselines while guaranteeing safety.}
\label{Tab:comparing with baselines_transposed}
\centering
\small 
\renewcommand{\arraystretch}{1.25} 
\setlength{\tabcolsep}{8pt} 

\resizebox{.8\textwidth}{!}{
\begin{tabular}{clccccc}
\toprule
\multicolumn{2}{c}{\multirow{2}{*}{Method}} & \multicolumn{4}{c}{PSM Tasks} & \multicolumn{1}{c}{Overall} \\ 
\cmidrule(lr){3-6} \cmidrule(lr){7-7}
 & & NeedleReach & GauzeRetrieve & NeedlePick & PegTransfer & Aggregate \\ 
\midrule

\multirow{2}{*}{\textbf{Reinforcement Learning}} 
 & SAC\citep{sac} & \textbf{1.00}\scriptsize{($\pm$.00)} & 0.00\scriptsize{($\pm$.00)} & 0.00\scriptsize{($\pm$.00)} & 0.00\scriptsize{($\pm$.00)} & \cellcolor[HTML]{EFEFEF}0.00\scriptsize{($\pm$.00)} \\
 & DDPG\citep{ddpg} & \textbf{1.00}\scriptsize{($\pm$.00)} & 0.00\scriptsize{($\pm$.00)} & 0.00\scriptsize{($\pm$.00)} & 0.00\scriptsize{($\pm$.00)} & \cellcolor[HTML]{EFEFEF}0.00\scriptsize{($\pm$.00)} \\
\cmidrule{1-7}

\multirow{4}{*}{\textbf{Imitation Learning}} 
 & BC\citep{bc} & \textbf{1.00}\scriptsize{($\pm$.00)} & 0.07\scriptsize{($\pm$.05)} & 0.21\scriptsize{($\pm$.06)} & 0.56\scriptsize{($\pm$.11)} & \cellcolor[HTML]{EFEFEF}0.40\scriptsize{($\pm$.05)} \\
 & SQIL\citep{sqil} & 0.07\scriptsize{($\pm$.09)} & 0.00\scriptsize{($\pm$.00)} & 0.00\scriptsize{($\pm$.00)} & 0.02\scriptsize{($\pm$.05)} & \cellcolor[HTML]{EFEFEF}0.00\scriptsize{($\pm$.00)} \\
 & VINN\citep{vinn} & 0.89\scriptsize{($\pm$.06)} & 0.01\scriptsize{($\pm$.02)} & 0.02\scriptsize{($\pm$.02)} & 0.05\scriptsize{($\pm$.04)} & \cellcolor[HTML]{EFEFEF}0.02\scriptsize{($\pm$.02)} \\
\cmidrule{1-7}

\multirow{5}{*}{\shortstack[c]{\textbf{Demonstration-guided}\\\textbf{Reinforcement Learning}}} 
 & DDPGBC\citep{ddpgher} & \textbf{1.00}\scriptsize{($\pm$.00)} & 0.63\scriptsize{($\pm$.11)} & 0.91\scriptsize{($\pm$.05)} & 0.48\scriptsize{($\pm$.22)} & \cellcolor[HTML]{EFEFEF}0.80\scriptsize{($\pm$.04)} \\
 & AMP\citep{amp} & 0.99\scriptsize{($\pm$.02)} & 0.00\scriptsize{($\pm$.00)} & 0.00\scriptsize{($\pm$.00)} & 0.00\scriptsize{($\pm$.00)} & \cellcolor[HTML]{EFEFEF}0.00\scriptsize{($\pm$.00)} \\
 & CoL\citep{col} & \textbf{1.00}\scriptsize{($\pm$.00)} & 0.71\scriptsize{($\pm$.16)} & \textbf{0.96}\scriptsize{($\pm$.05)} & 0.58\scriptsize{($\pm$.23)} & \cellcolor[HTML]{EFEFEF}0.85\scriptsize{($\pm$.06)} \\
 & AWAC\citep{awac} & 0.94\scriptsize{($\pm$.20)} & 0.43\scriptsize{($\pm$.43)} & 0.26\scriptsize{($\pm$.33)} & 0.31\scriptsize{($\pm$.32)} & \cellcolor[HTML]{EFEFEF}0.46\scriptsize{($\pm$.19)} \\
 & DEX\cite{dex} & \textbf{1.00}\scriptsize{($\pm$.00)} & 0.73\scriptsize{($\pm$.12)} & 0.94\scriptsize{($\pm$.05)} & 0.73\scriptsize{($\pm$.20)} & \cellcolor[HTML]{EFEFEF}\textbf{0.89}\scriptsize{($\pm$.03)} \\
\cmidrule{1-7}

\multirow{3}{*}{\shortstack[c]{\textbf{SSP (Proposed)}}} 
 & SSP-DP & 0.48\scriptsize{($\pm$.05)} & 0.24\scriptsize{($\pm$.06)} & 0.12\scriptsize{($\pm$.03)} & \textbf{0.81}\scriptsize{($\pm$.19)} & \cellcolor[HTML]{EFEFEF}0.41\scriptsize{($\pm$.08)} \\
 & SSP-CLF & 0.63\scriptsize{($\pm$.37)} & \textbf{0.99}\scriptsize{($\pm$.02)} & 0.87\scriptsize{($\pm$.06)} & 0.21\scriptsize{($\pm$.05)} & \cellcolor[HTML]{EFEFEF}0.68\scriptsize{($\pm$.30)} \\
 & SSP-DEX & \textbf{1.00}\scriptsize{($\pm$.00)} & 0.75\scriptsize{($\pm$.08)} & 0.74\scriptsize{($\pm$.16)} & 0.60\scriptsize{($\pm$.09)} & \cellcolor[HTML]{EFEFEF}0.78\scriptsize{($\pm$.14)} \\
\bottomrule
\end{tabular}
}
\end{table*}

\noindent \textbf{Evaluation Metrics} We report several key metrics evaluating the performance of all the methods in \Cref{tab:success_rate}:

\noindent - \textit{Success Rate with Violation ($\%$)}: The percentage of trials where the primary task is completed no matter collision happen or not.
This is the original measure of the performance of the base policy.

\noindent - \textit{Success Rate w/o Violation ($\%$)}: The percentage of trials where the primary task is completed and no collision with an no-go zone occurs.
This is the ultimate measure of the performance of our system.

\noindent - \textit{Collision Rate ($\%$)}: The percentage of trials where the robot end-effector makes contact with a no-go zone.
This specifically isolates the safety performance.

\noindent - \textit{Inference time (ms)}: This measures the average inference time required for execute a single trajectory, indicating that including CLF or CBF in the system does not introduce significant latency.

\noindent - \textit{Safe Margin}: Safety margin is used as a quantitative measure to evaluate the performance of the CBF-based safety controller.
We record the smallest $b(x)$ along the trajectory.
A positive margin indicates that the system remains within the safe set, while a smaller or negative margin reflects proximity to, or violation of, the safety boundary.

\noindent - \textit{Uncertainty Quantification}: We quantify the uncertainty (\ref{eqn:un1})(\ref{eqn:un2}) in the dynamics model learned with a Neural ODE.

\noindent \textbf{Results} The quantitative results in \Cref{tab:success_rate} demonstrate the critical efficacy of the proposed framework in enforcing safety without severely compromising task performance.
A comparison between baseline policies (DEX, CLF, DP) and our framework instantiated with these policy (SSP-) reveals a stark contrast in collision rates.
Unconstrained policies frequently fail to avoid no-go zones, evidenced by Collision Rates often reaching 100\% in tightly constrained environments like NeedlePick.
Conversely, the integration of the robust CBF safety filter consistently reduces the collision rate to near zero (e.g., SSP-DEX achieves a 0.00 collision rate in NeedlePick-Cylinder compared to 1.00 for DEX).
Consequently, the Success Rate without Violation, the ultimate measure of safe autonomy, improves dramatically.
Furthermore, the Safe Margin metric validates the mathematical robustness of the approach.
While baseline methods exhibit negative margins (indicating safety violations), the SSP methods maintain positive margins, quantitatively confirming that the system strictly adheres to the defined safety boundaries.
Notably, this safety assurance incurs negligible computational overhead; the Inference Time increases only marginally, ensuring the framework remains sufficiently fast for real-time surgical control.

Next, we compare the performance of baseline methods (in no-go zone-free environments) against our method (in the same environments but with no-go zones), as shown in \Cref{Tab:comparing with baselines_transposed}. 
When instantiated with DEX as the high-level policy, our framework allows safety guarantees without a significant drop in performance.
Moreover, we visualize one trajectory and record the safe margin along the trajectory, as shown in \Cref{fig:trajectories_visualization}, for the task NeedlePick-Cylinder solved SSP-CLF .

\subsection{Real World Experiments}

\begin{figure*}[t]
    \centering
    \setlength{\tabcolsep}{2pt} 
    \begin{tabular}{@{} c c c c c c @{}}
        
        \raisebox{-0.5\height}{(a)} & 
        \raisebox{-0.5\height}{\includegraphics[width=0.19\linewidth]{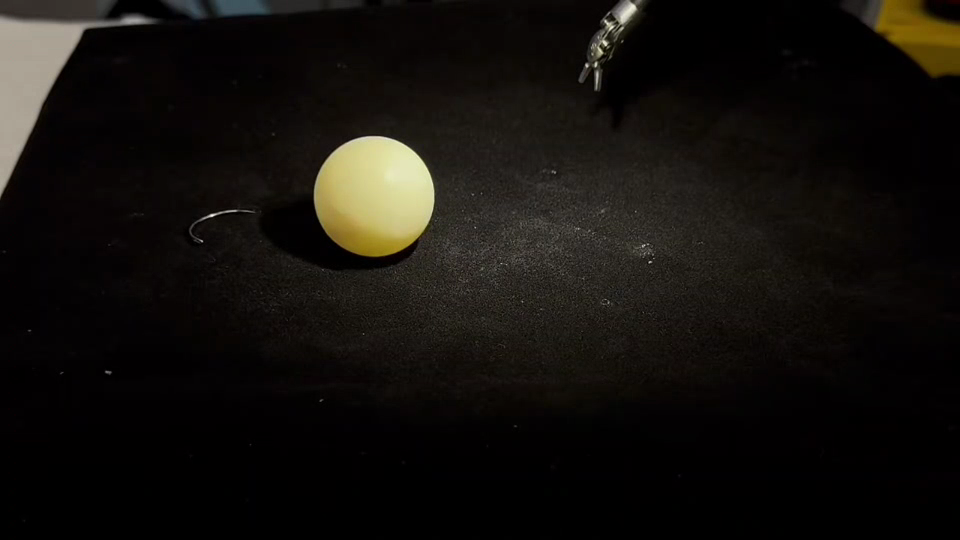}} & 
        \raisebox{-0.5\height}{\includegraphics[width=0.19\linewidth]{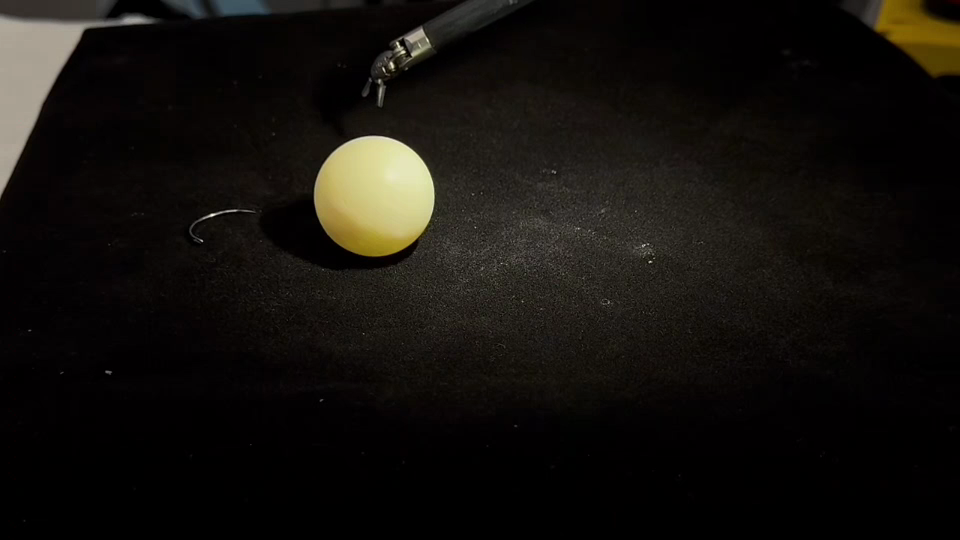}} & 
        \raisebox{-0.5\height}{\includegraphics[width=0.19\linewidth]{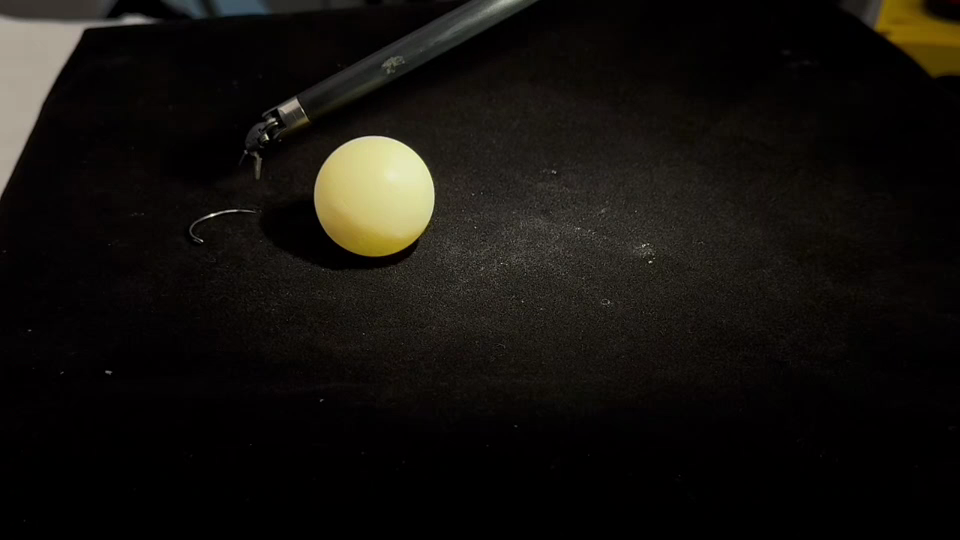}} & 
        \raisebox{-0.5\height}{\includegraphics[width=0.19\linewidth]{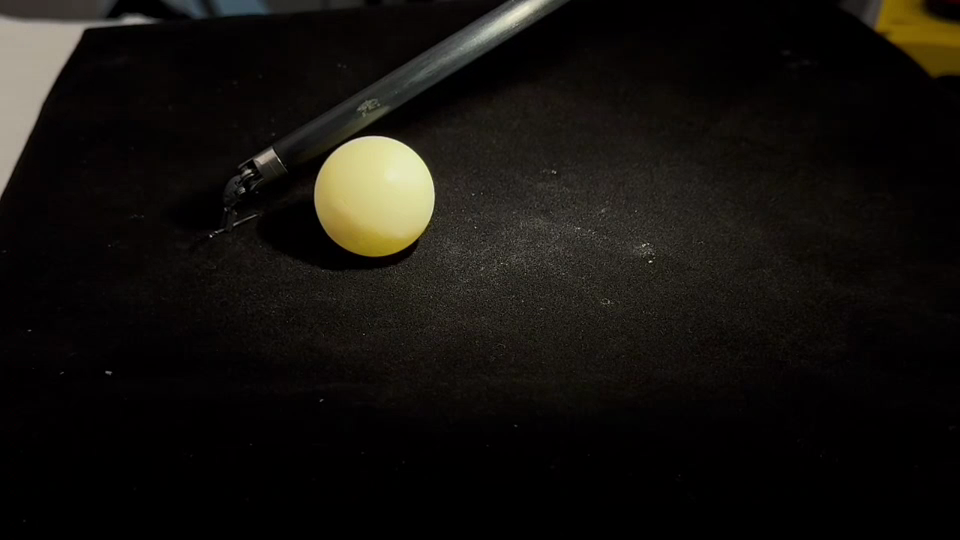}} & 
        \raisebox{-0.5\height}{\includegraphics[width=0.19\linewidth]{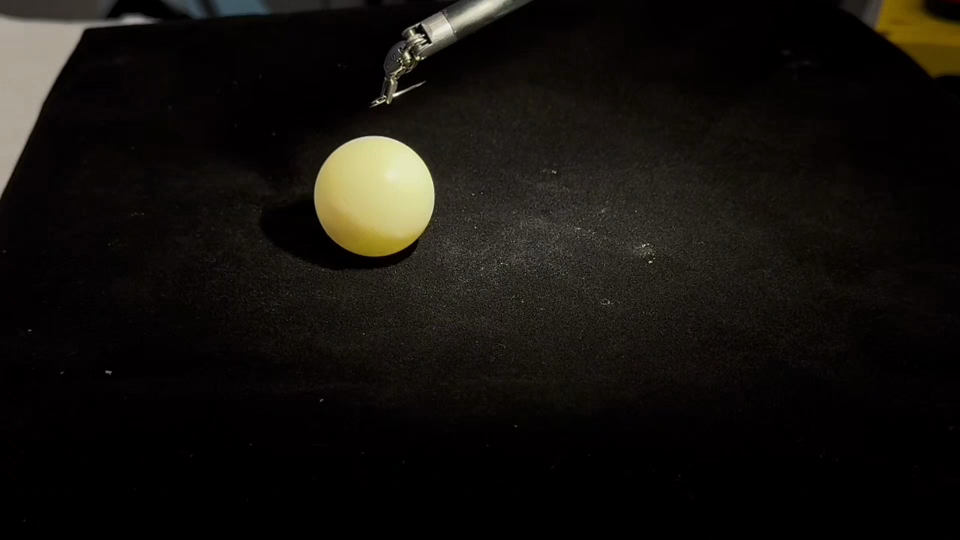}} \\
        \addlinespace
        
        \raisebox{-0.5\height}{(b)} & 
        \raisebox{-0.5\height}{\includegraphics[width=0.19\linewidth]{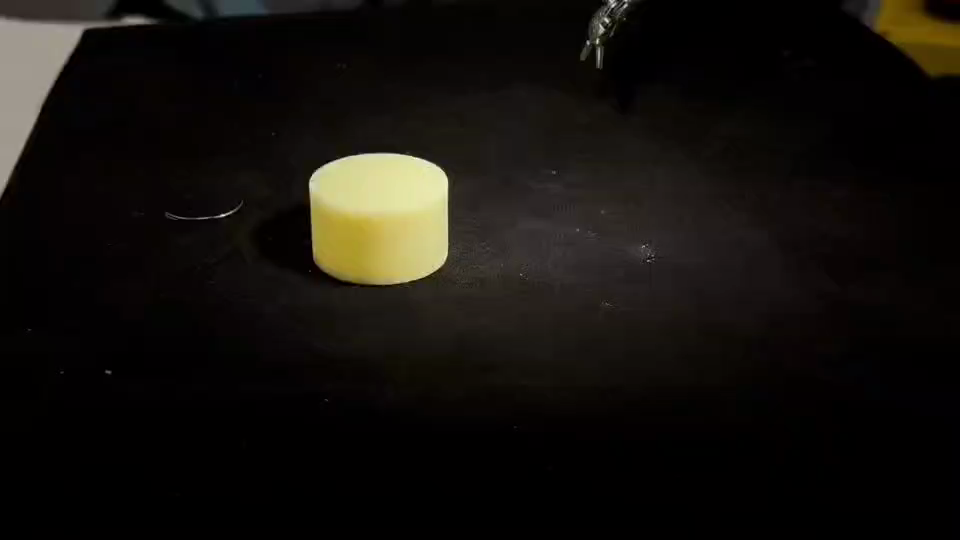}} & 
        \raisebox{-0.5\height}{\includegraphics[width=0.19\linewidth]{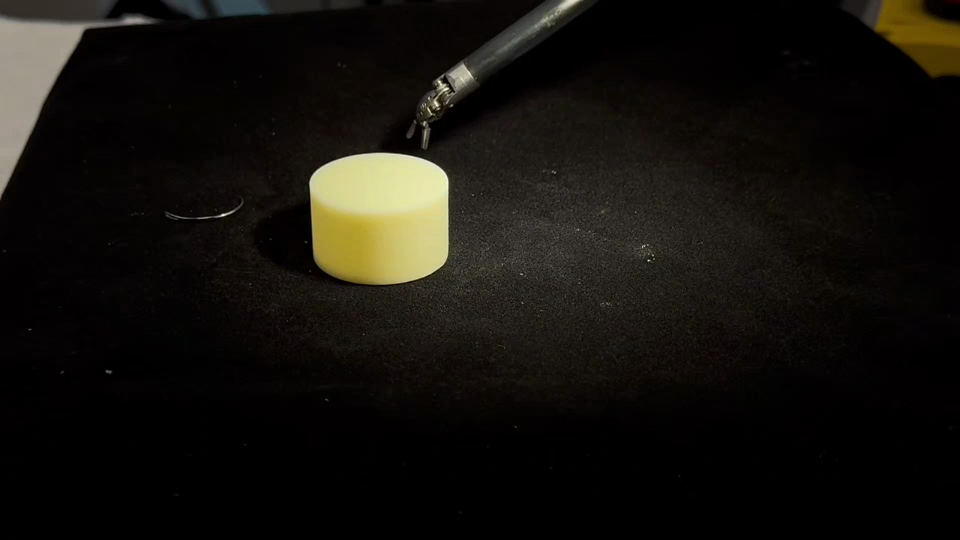}} & 
        \raisebox{-0.5\height}{\includegraphics[width=0.19\linewidth]{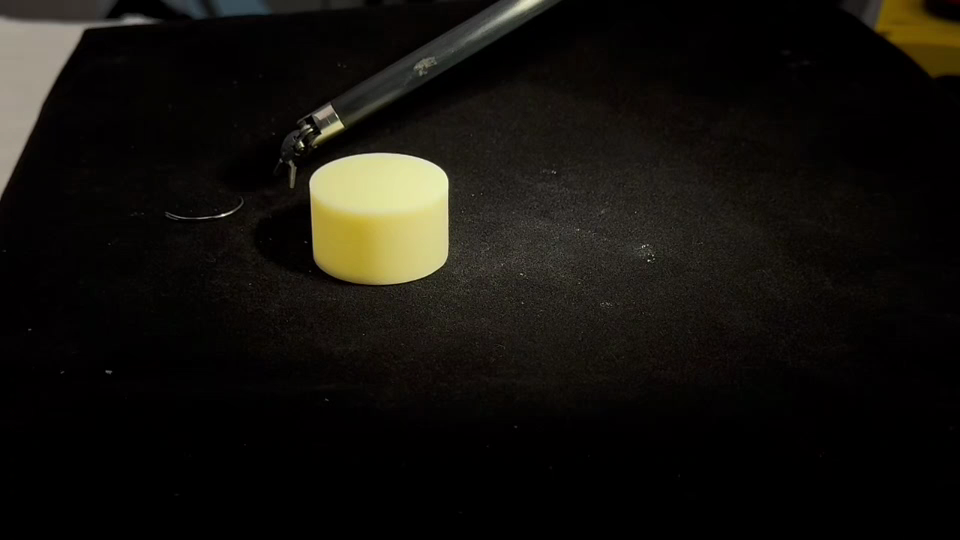}} & 
        \raisebox{-0.5\height}{\includegraphics[width=0.19\linewidth]{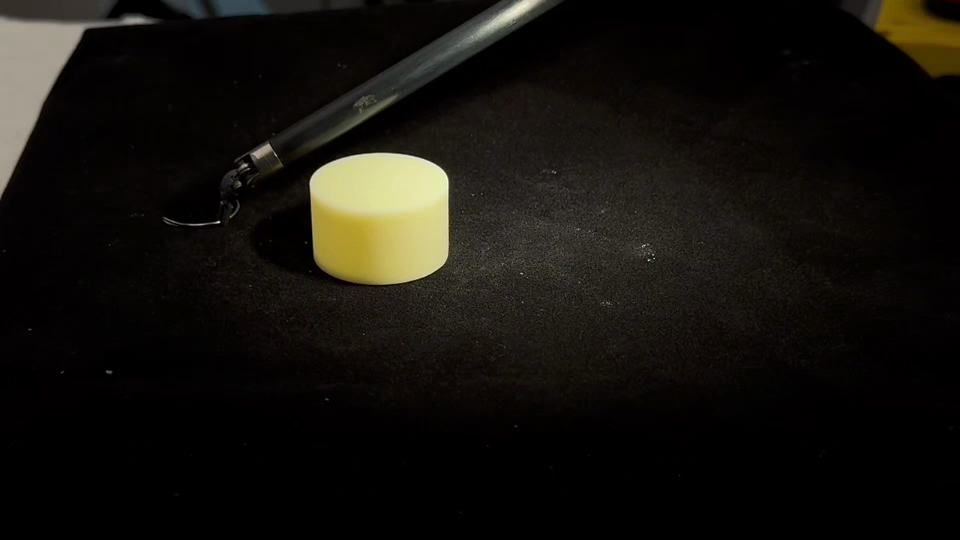}} & 
        \raisebox{-0.5\height}{\includegraphics[width=0.19\linewidth]{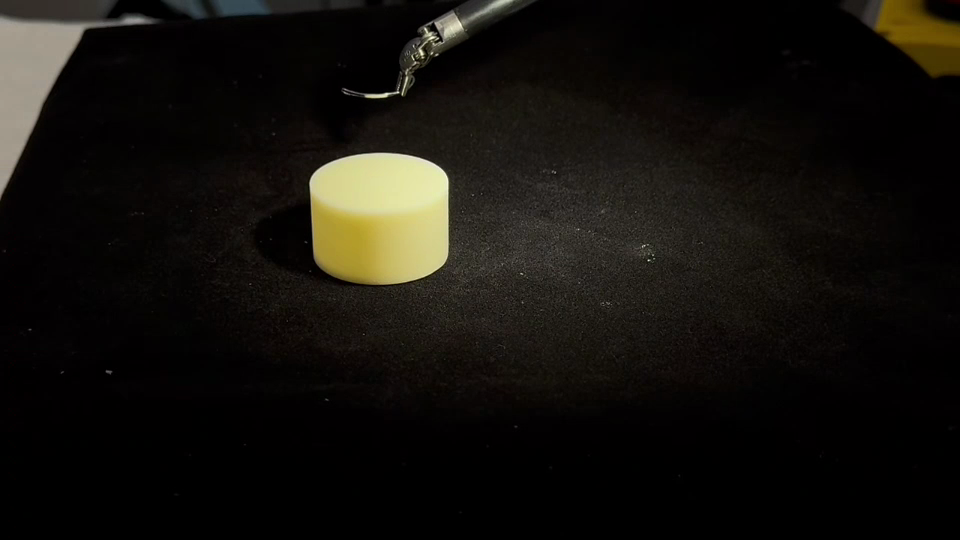}} \\
        \addlinespace
        
        \raisebox{-0.5\height}{(c)} & 
        \raisebox{-0.5\height}{\includegraphics[width=0.19\linewidth]{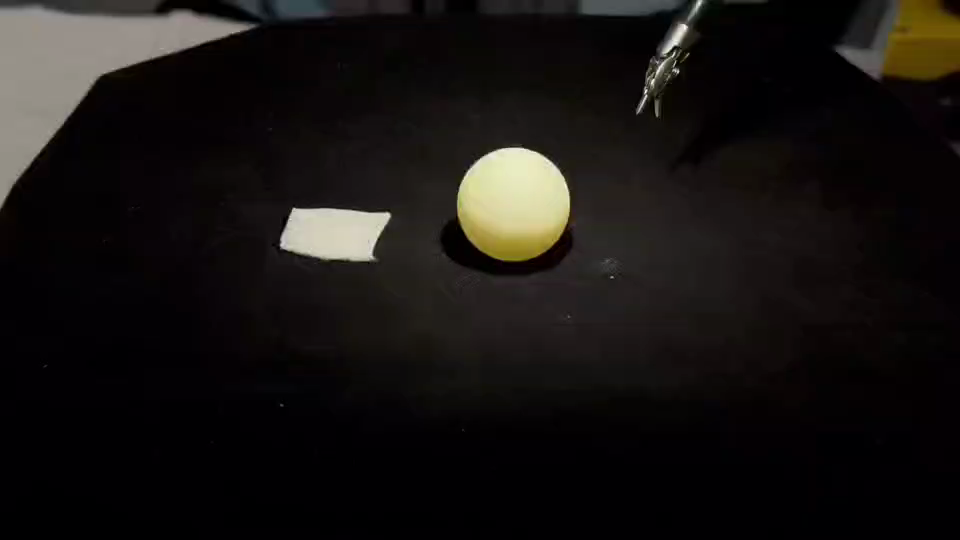}} & 
        \raisebox{-0.5\height}{\includegraphics[width=0.19\linewidth]{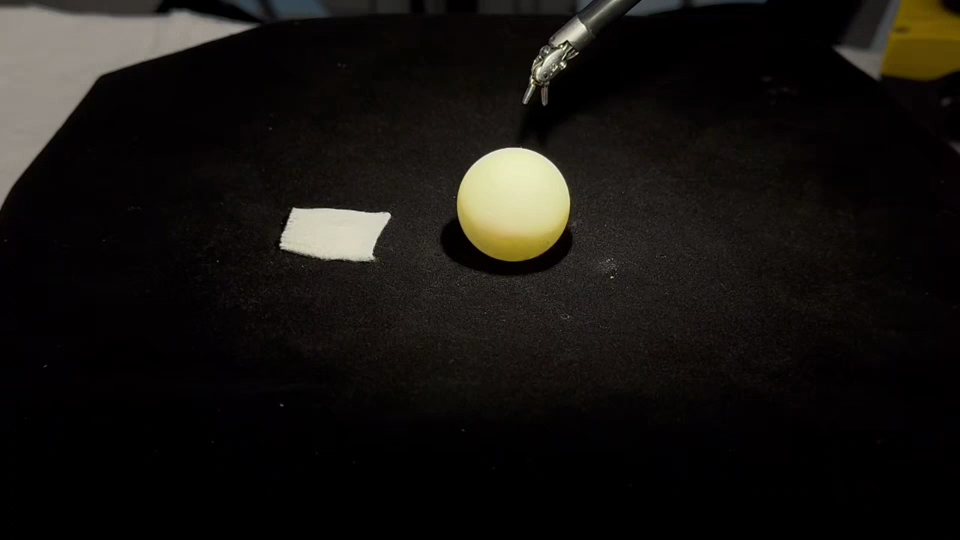}} & 
        \raisebox{-0.5\height}{\includegraphics[width=0.19\linewidth]{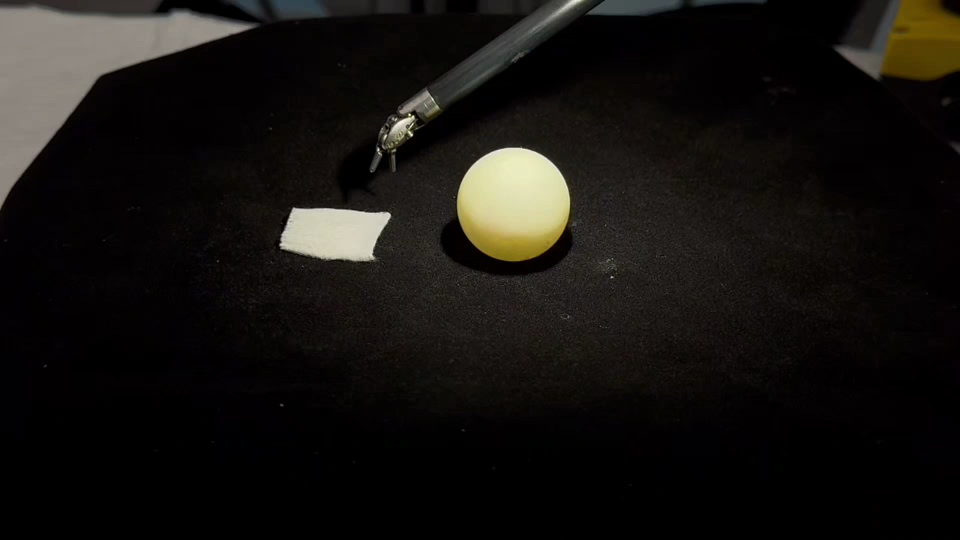}} & 
        \raisebox{-0.5\height}{\includegraphics[width=0.19\linewidth]{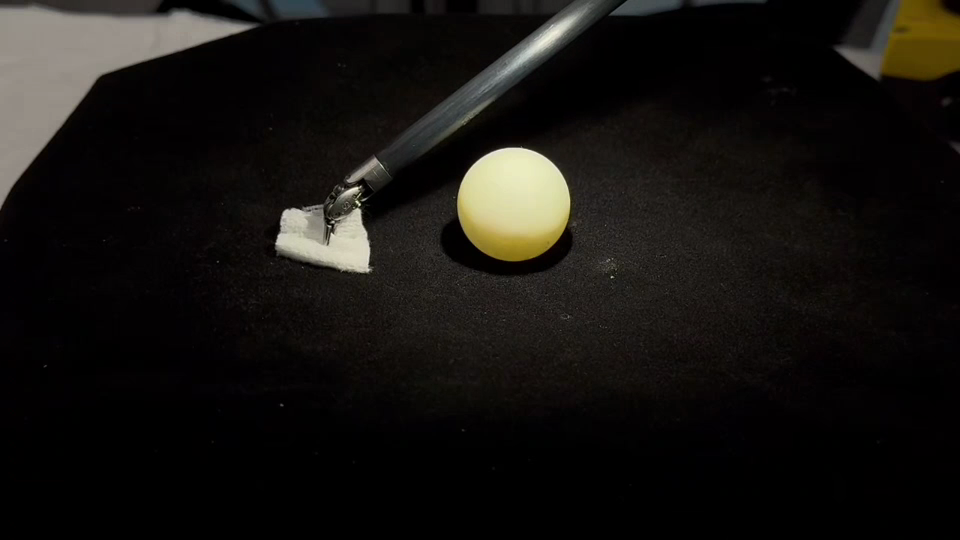}} & 
        \raisebox{-0.5\height}{\includegraphics[width=0.19\linewidth]{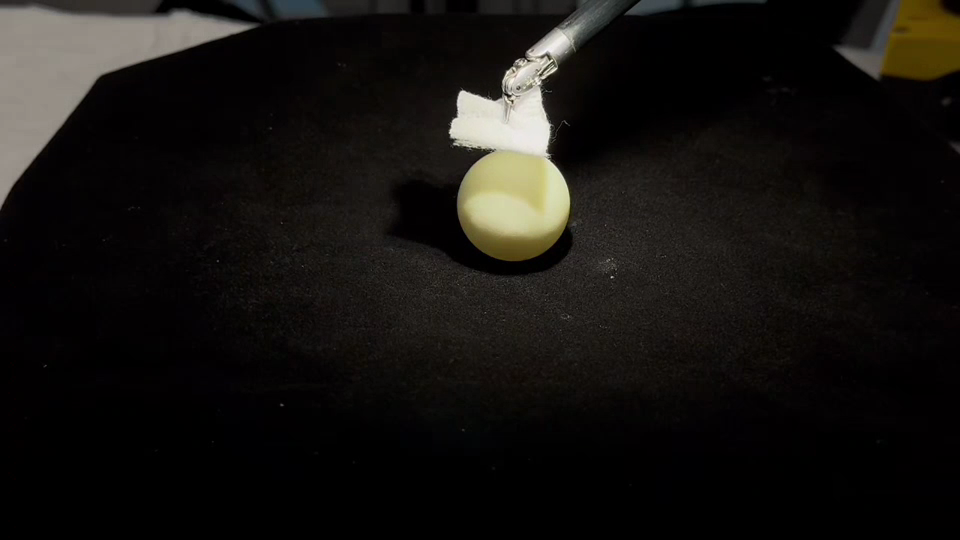}} \\
        \addlinespace
        
        \raisebox{-0.5\height}{(d)} & 
        \raisebox{-0.5\height}{\includegraphics[width=0.19\linewidth]{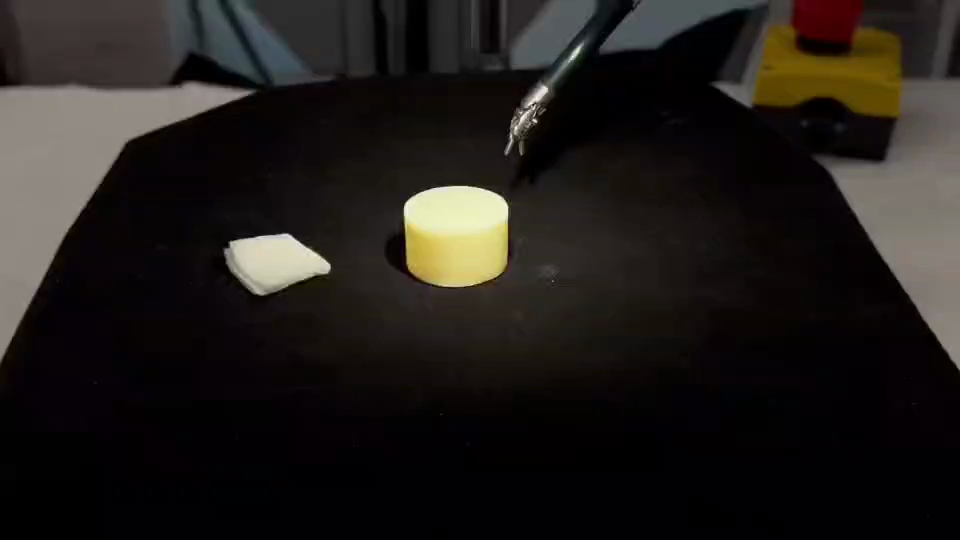}} & 
        \raisebox{-0.5\height}{\includegraphics[width=0.19\linewidth]{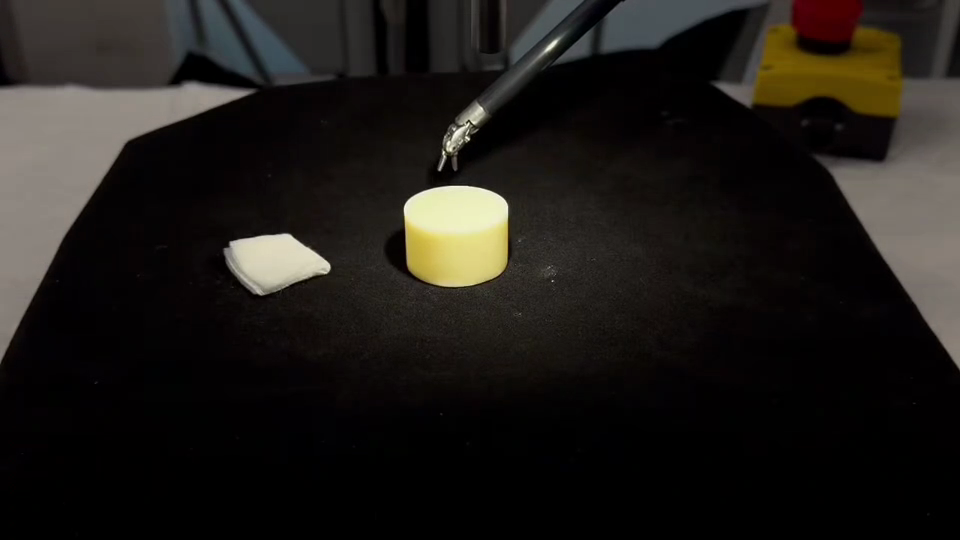}} & 
        \raisebox{-0.5\height}{\includegraphics[width=0.19\linewidth]{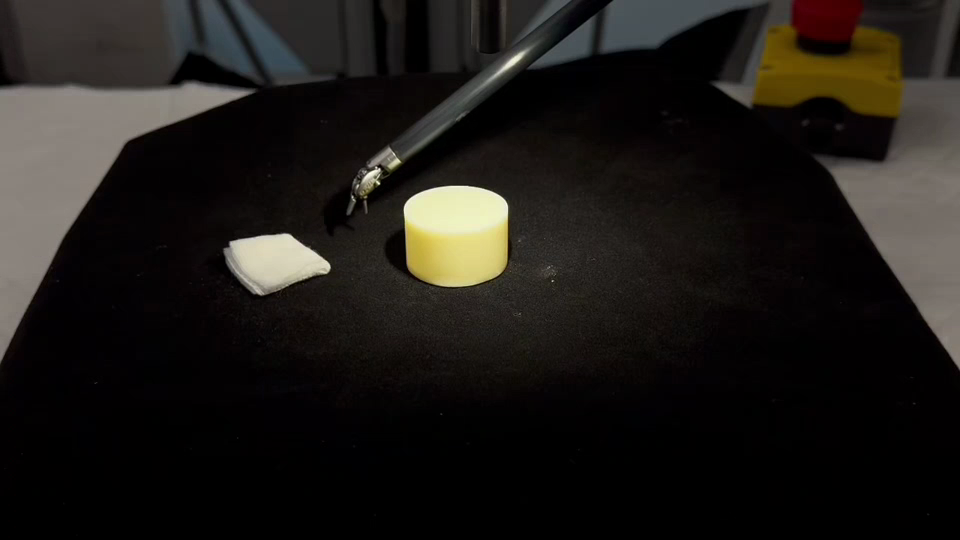}} & 
        \raisebox{-0.5\height}{\includegraphics[width=0.19\linewidth]{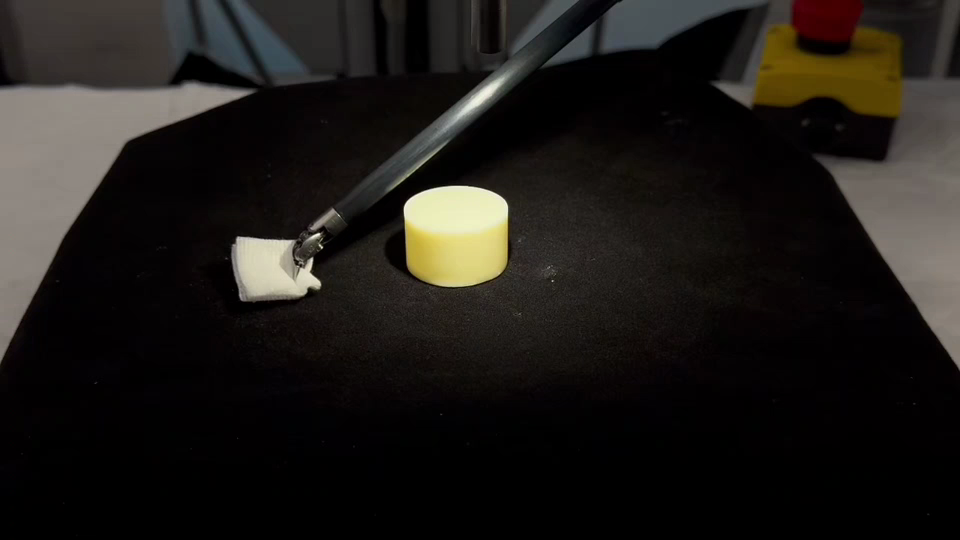}} & 
        \raisebox{-0.5\height}{\includegraphics[width=0.19\linewidth]{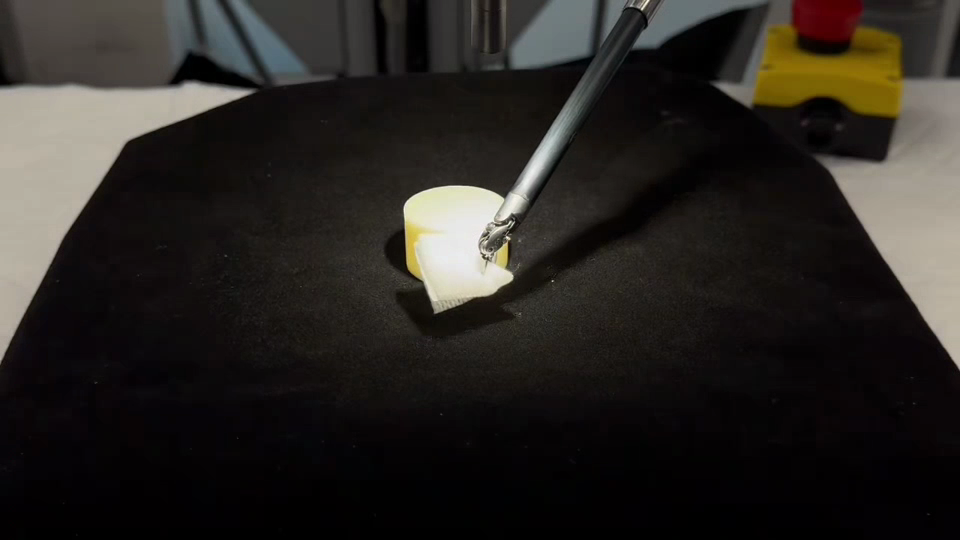}} \\
        
    \end{tabular}
    
    \vspace{0.5em}
    \caption{\textbf{Real World Experiments of Needle and Gauze Picking with No-Go Zones:} Sampled frames evaluating RL with safety constraints guaranteed by CBF. The rows show (a) needle pick with a sphere, (b) needle pick with a cylinder, (c) gauze retrieve with a sphere, and (d) gauze retrieve with a cylinder.}
    \label{fig:real_world_needle_gauze}
\end{figure*}

\begin{figure*}[t]
    \centering
    \setlength{\tabcolsep}{2pt} 
    \begin{tabular}{@{} c c c c c c @{}}
        
        \raisebox{-0.5\height}{(a)} & 
        \raisebox{-0.5\height}{\includegraphics[width=0.19\linewidth]{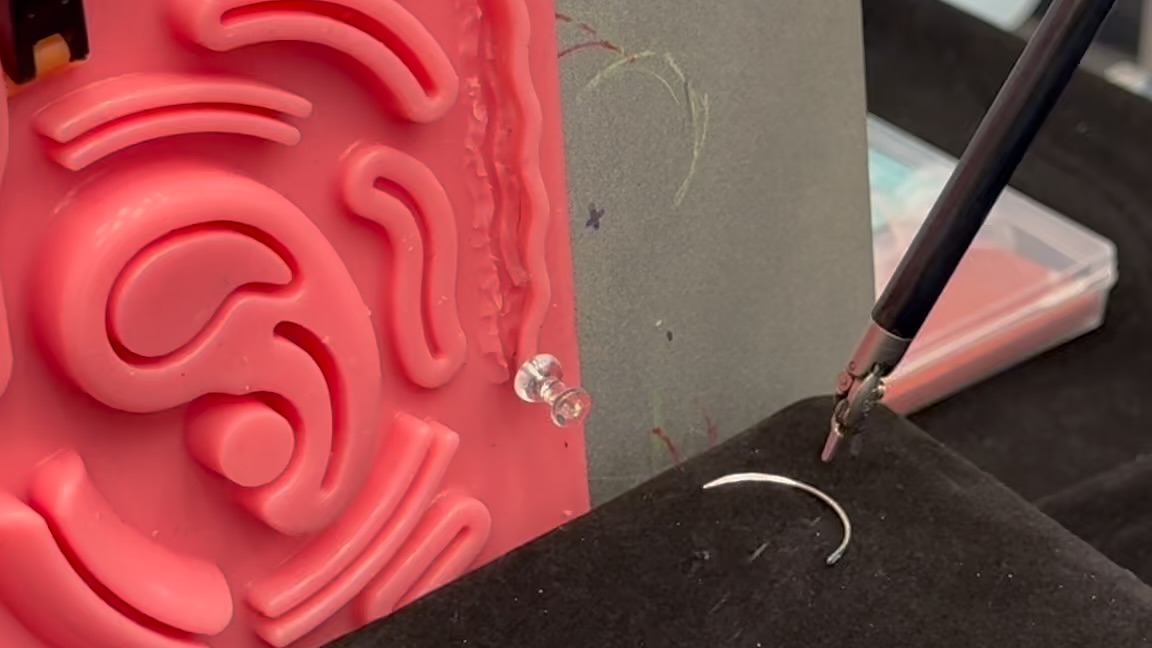}} & 
        \raisebox{-0.5\height}{\includegraphics[width=0.19\linewidth]{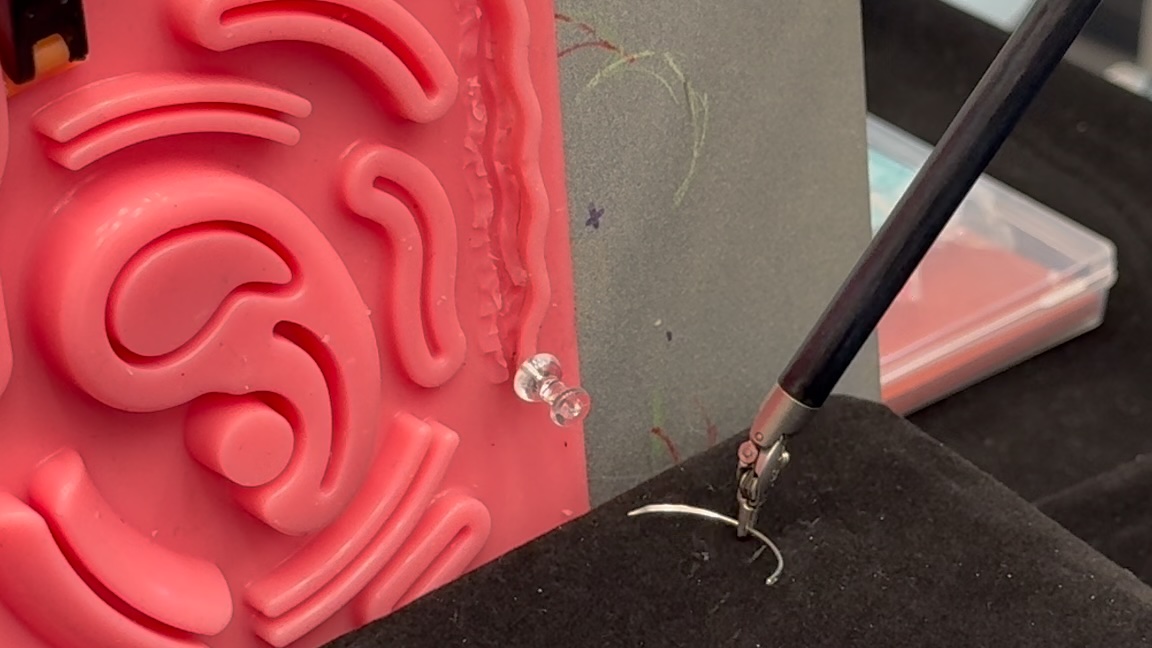}} & 
        \raisebox{-0.5\height}{\includegraphics[width=0.19\linewidth]{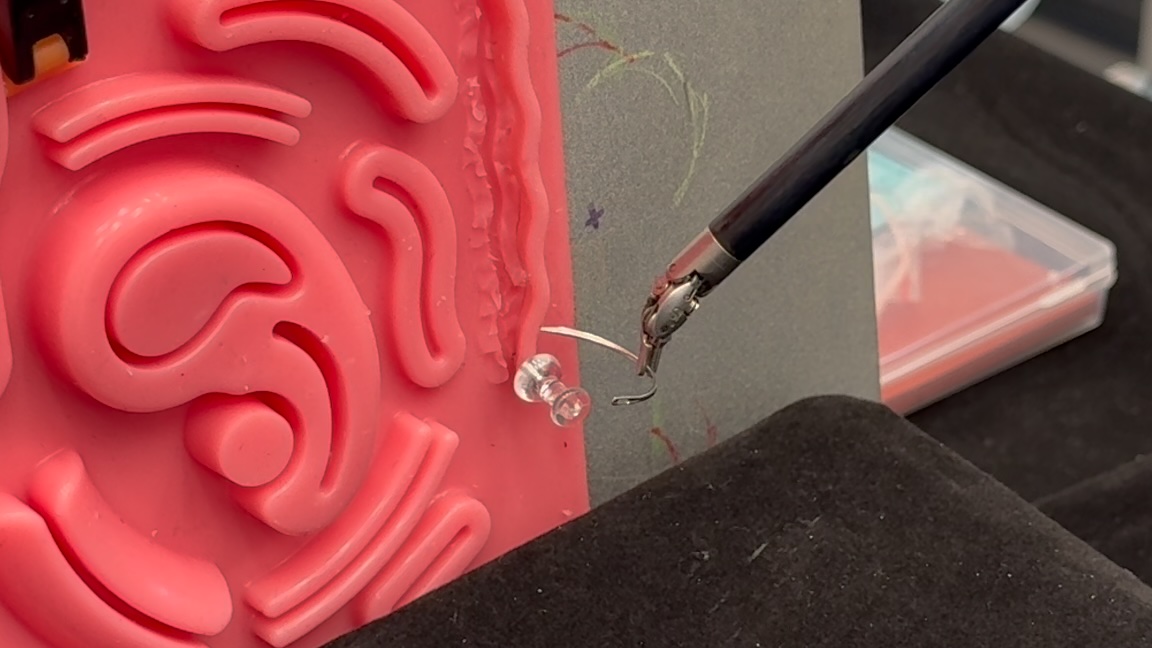}} & 
        \raisebox{-0.5\height}{\includegraphics[width=0.19\linewidth]{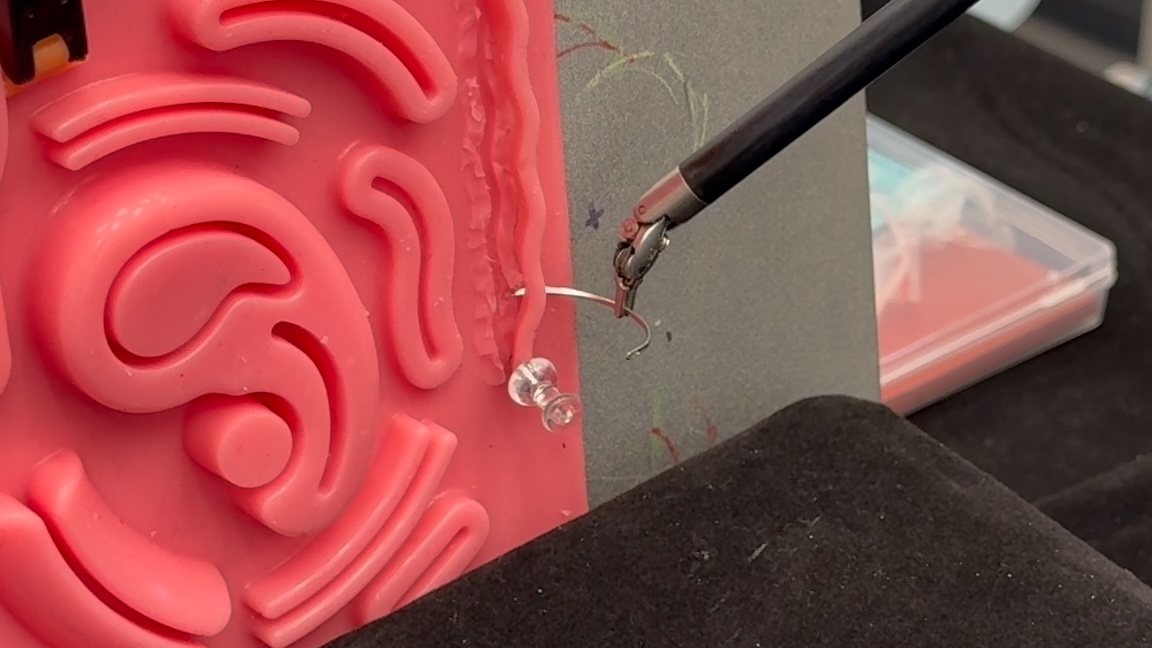}} & 
        \raisebox{-0.5\height}{\includegraphics[width=0.19\linewidth]{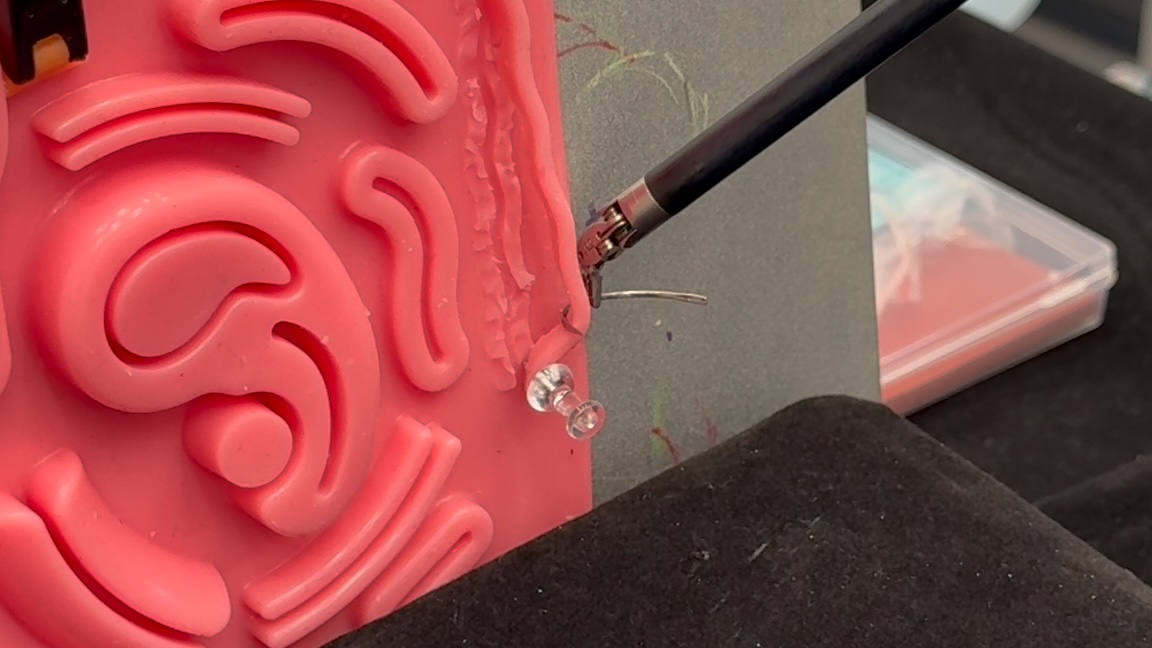}} \\
        \addlinespace
        
        \raisebox{-0.5\height}{(b)} & 
        \raisebox{-0.5\height}{\includegraphics[width=0.19\linewidth]{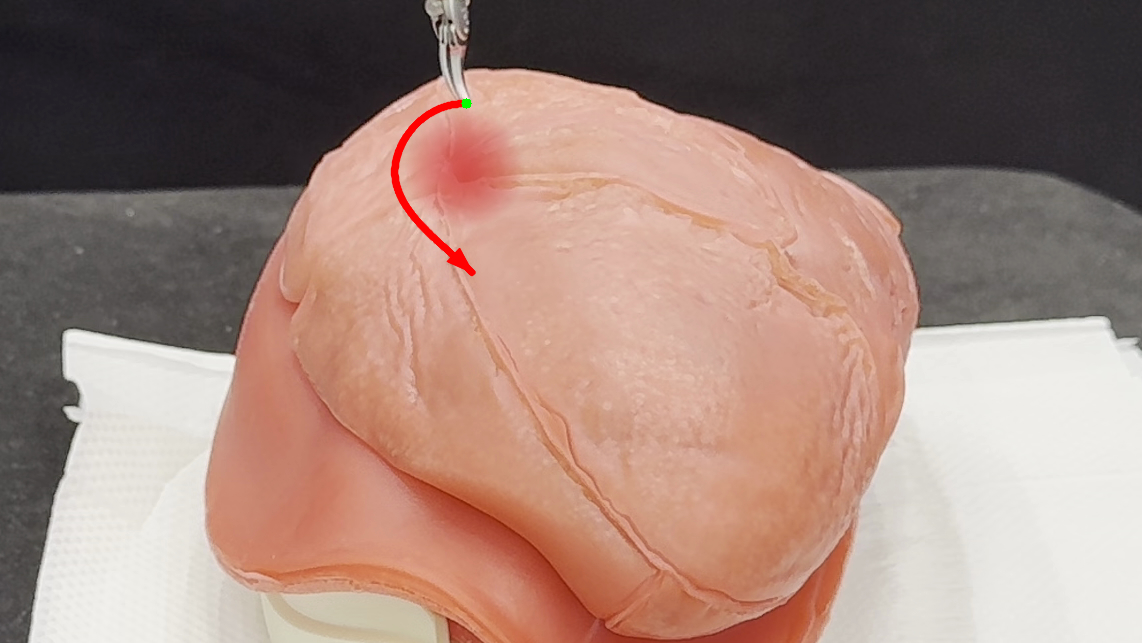}} & 
        \raisebox{-0.5\height}{\includegraphics[width=0.19\linewidth]{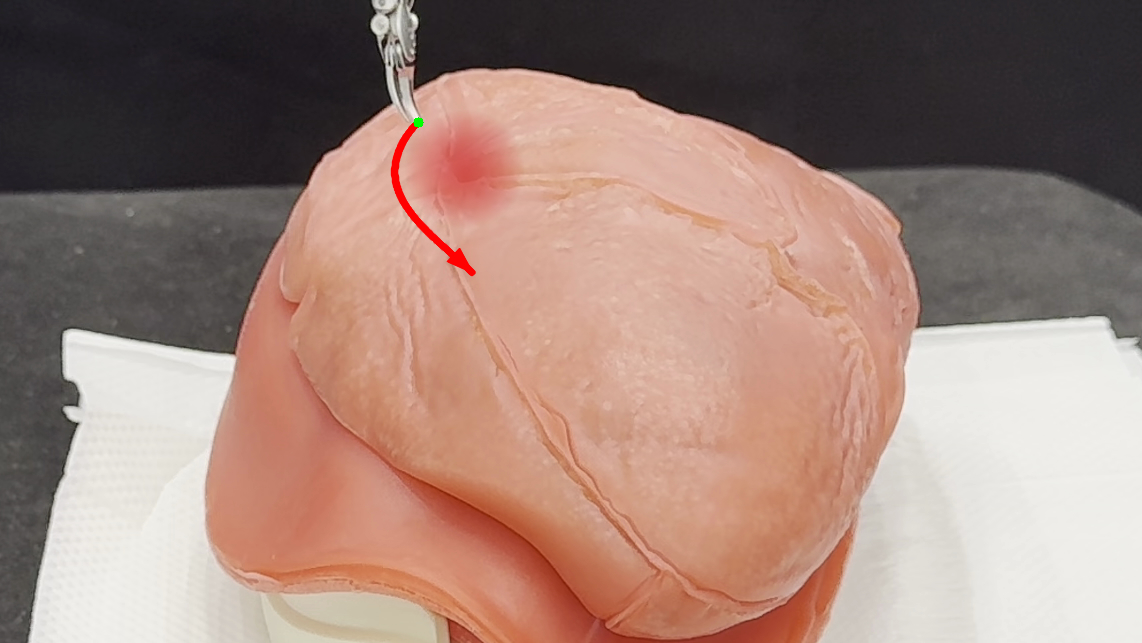}} & 
        \raisebox{-0.5\height}{\includegraphics[width=0.19\linewidth]{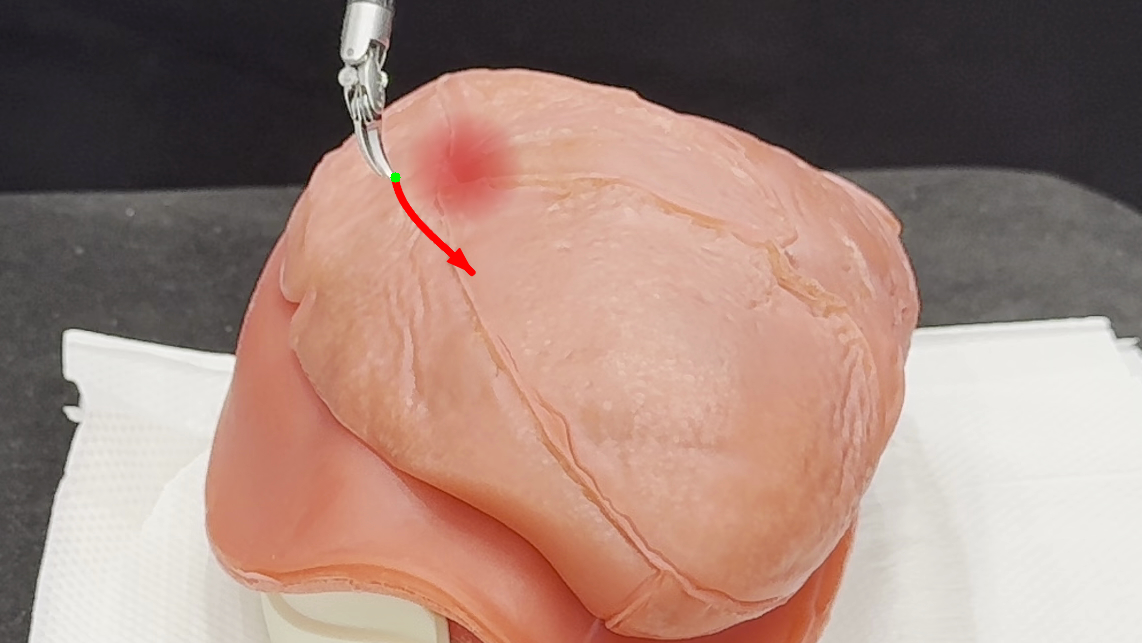}} & 
        \raisebox{-0.5\height}{\includegraphics[width=0.19\linewidth]{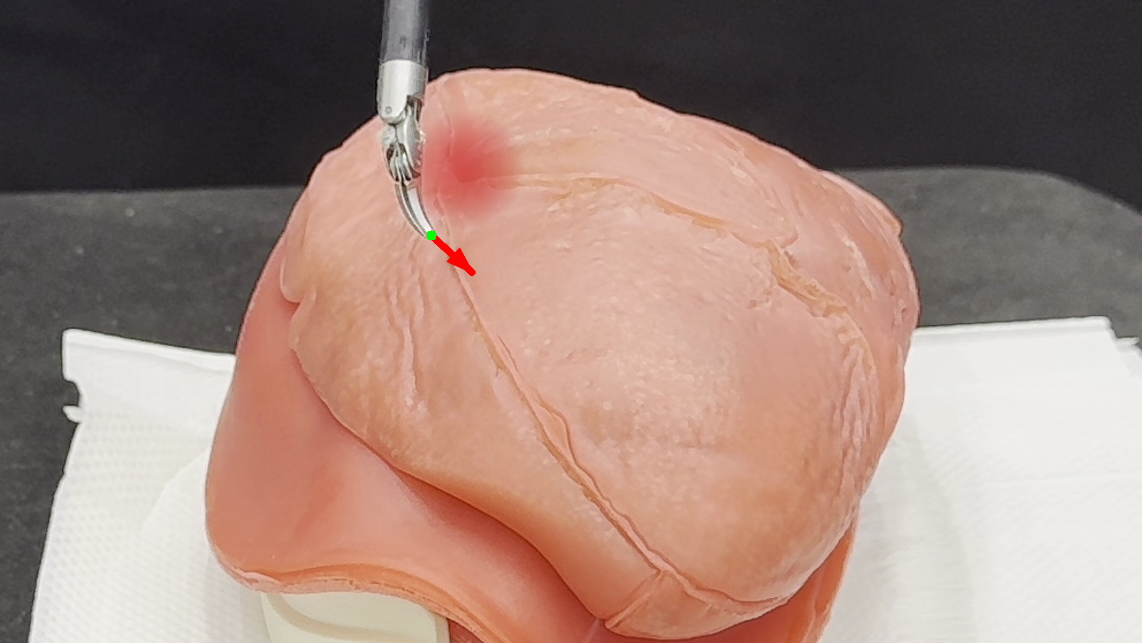}} & 
        \raisebox{-0.5\height}{\includegraphics[width=0.19\linewidth]{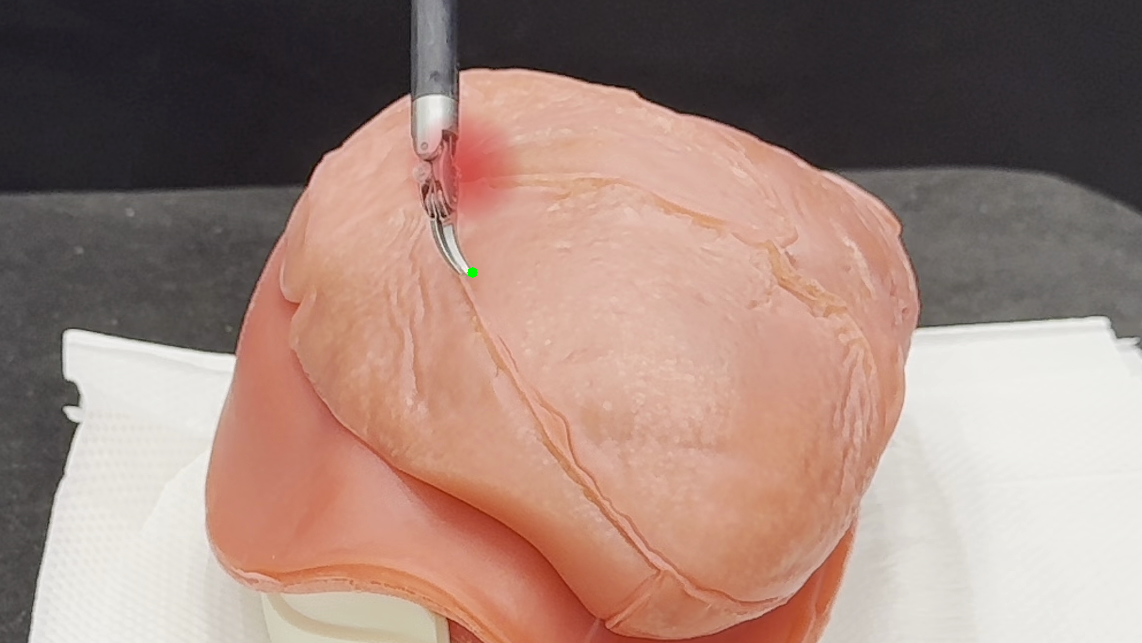}} \\
        \addlinespace
        
        \raisebox{-0.5\height}{(c)} & 
        \raisebox{-0.5\height}{\includegraphics[width=0.19\linewidth]{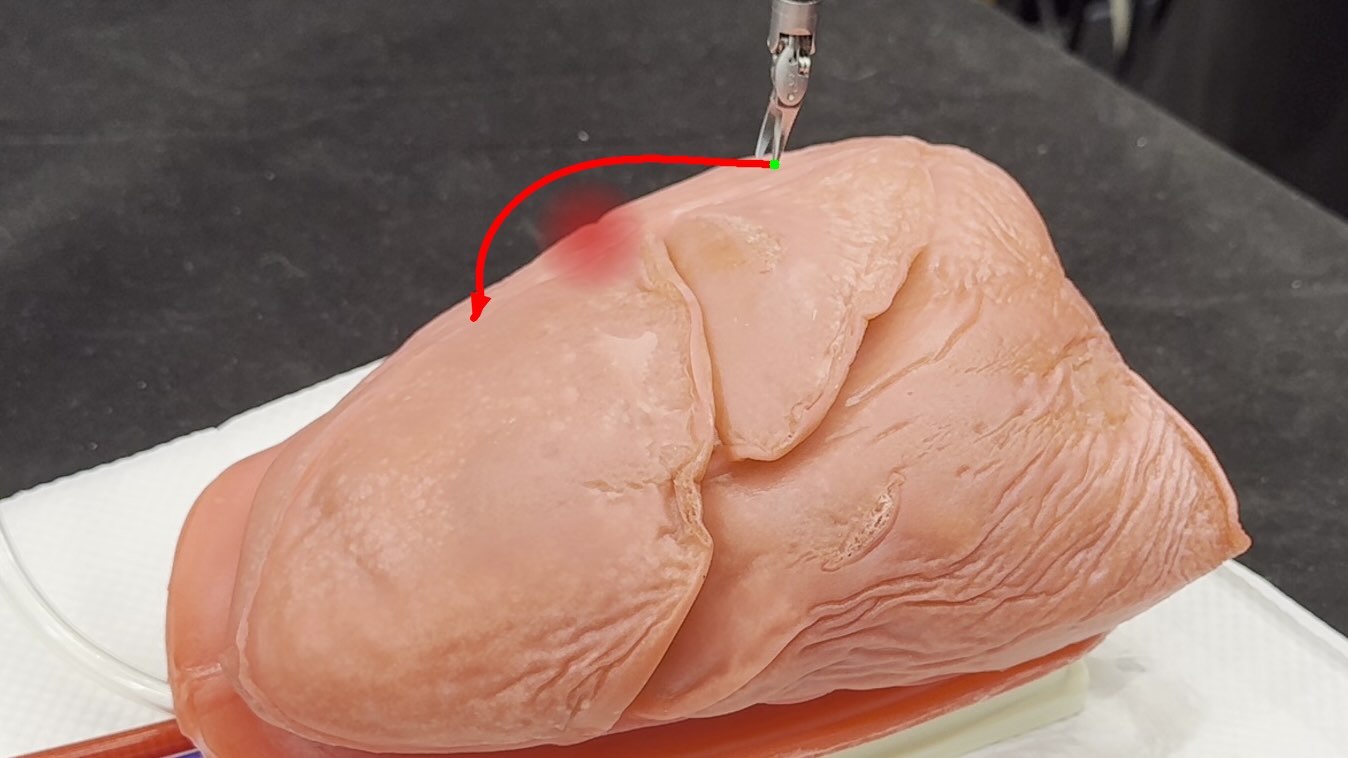}} & 
        \raisebox{-0.5\height}{\includegraphics[width=0.19\linewidth]{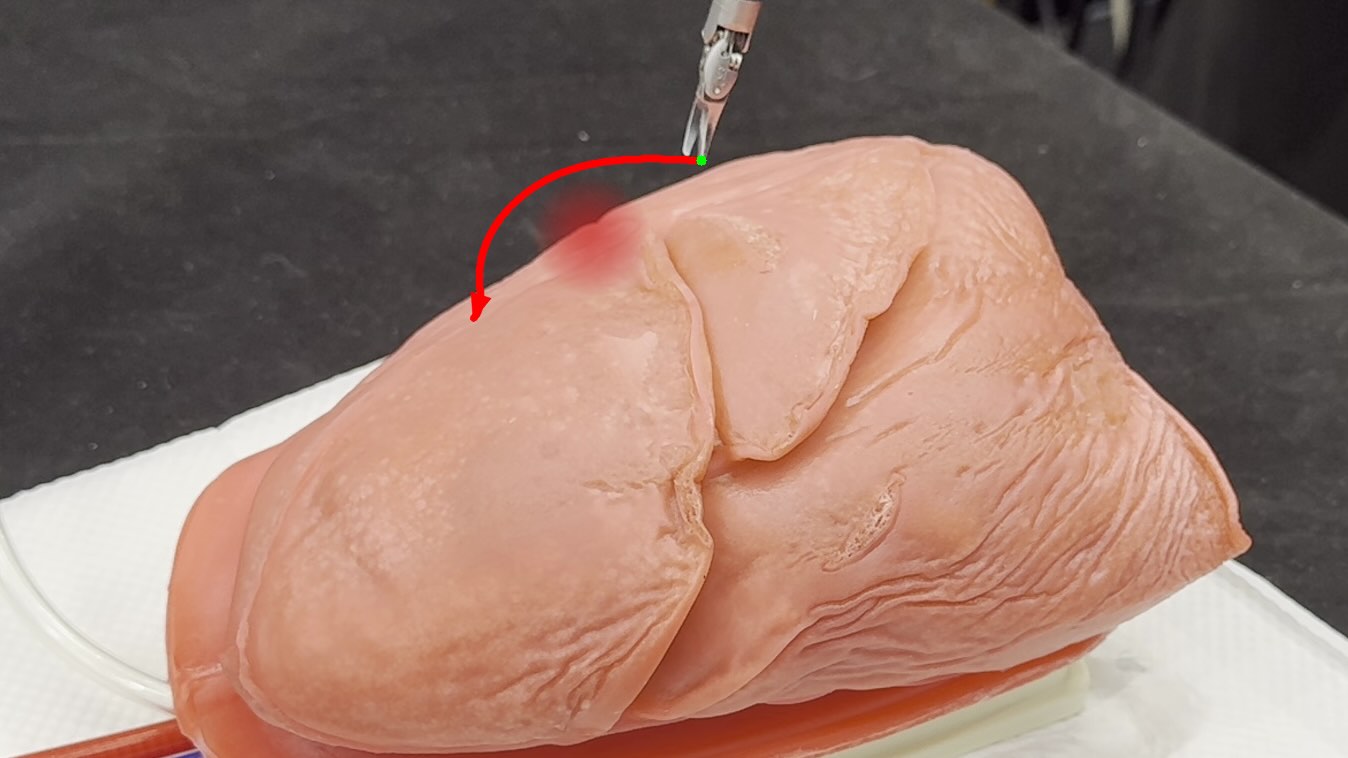}} & 
        \raisebox{-0.5\height}{\includegraphics[width=0.19\linewidth]{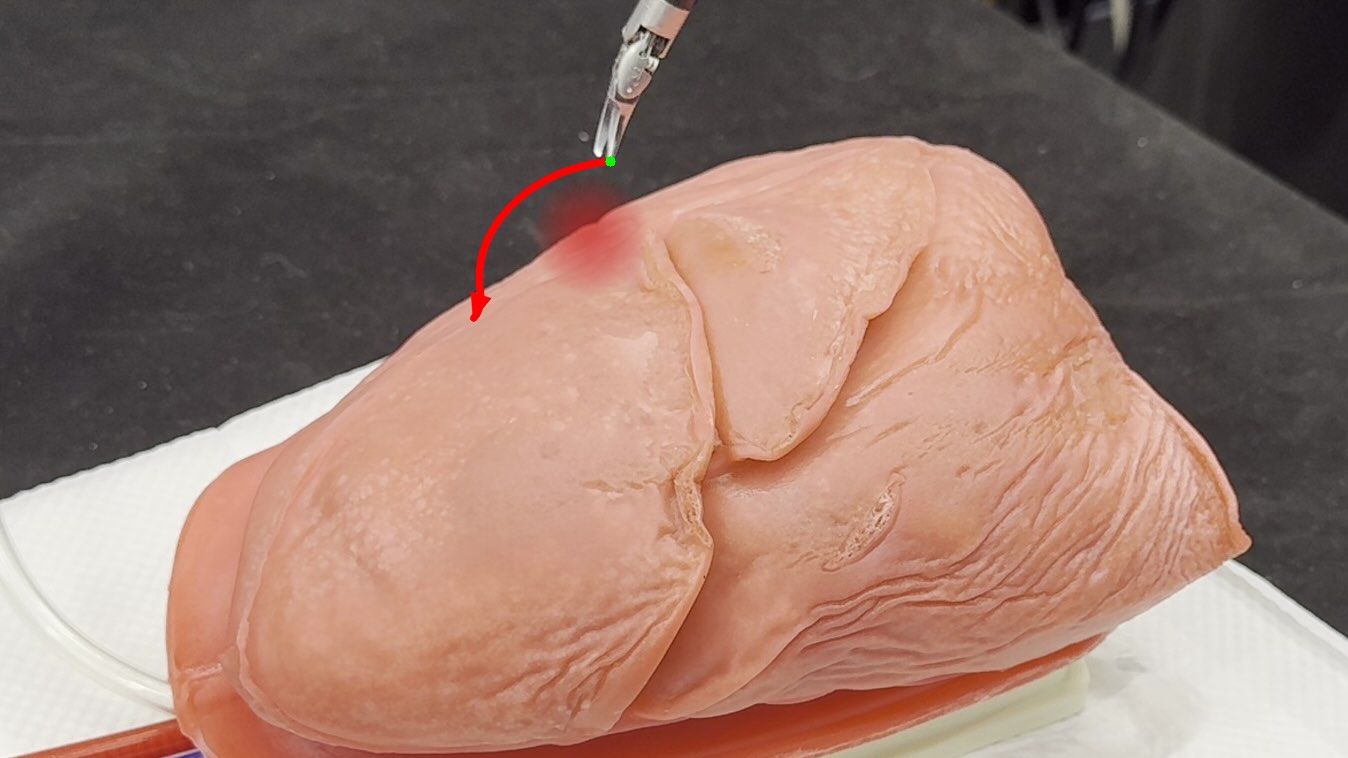}} & 
        \raisebox{-0.5\height}{\includegraphics[width=0.19\linewidth]{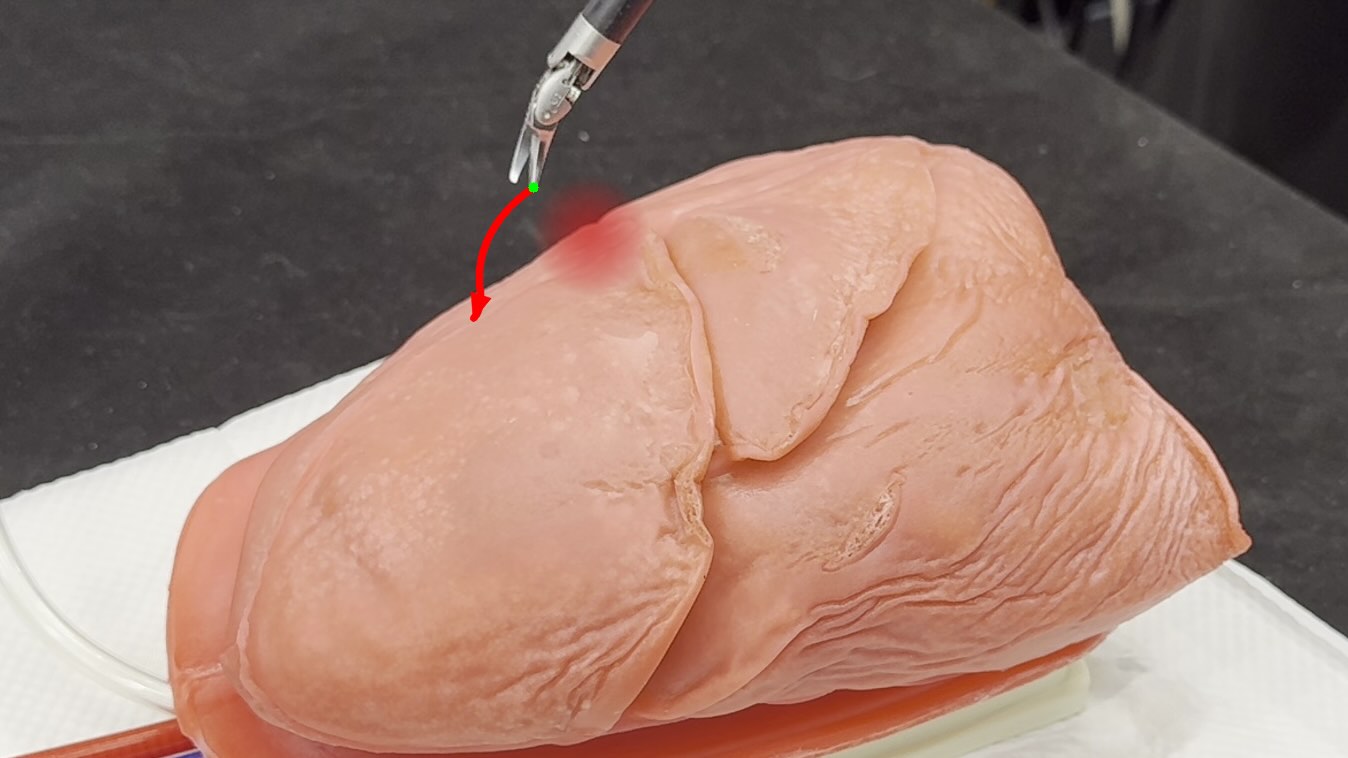}} & 
        \raisebox{-0.5\height}{\includegraphics[width=0.19\linewidth]{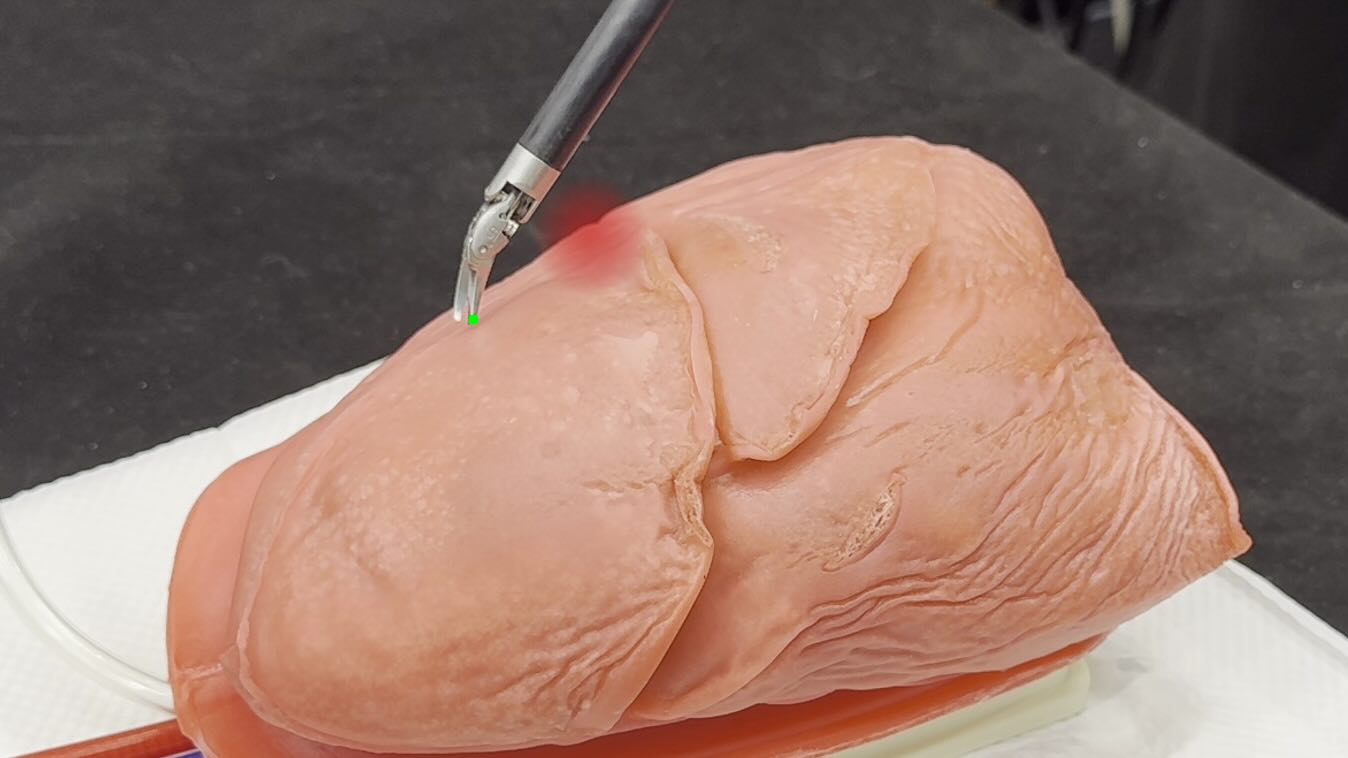}} \\
        
    \end{tabular}
    
    \vspace{0.5em}
    \caption{
    \textbf{Real World Experiments on dVRK.} 
    \textbf{(a)} Multi stage suturing sequence (RL + CLF). The task is decomposed into an RL based grasp phase for robust needle acquisition, followed by a CLF based insertion phase. 
    \textbf{(b) and (c)} Lung tumor resection task with safety constraints (Top and Side views, CLF + CBF). The CLF generates a nominal cutting trajectory, while the CBF enforces a hard safety constraint to prevent penetration into a spherical vascular safety region ($r=90$~mm). The controller smoothly deviates to avoid the no-go zone and converges back to the path.
    }
    \label{fig:real_world_experiments_suturing_and_lung}
\end{figure*}
To validate the practical efficacy and robustness of our framework, we conducted a series of experiments on a real da Vinci Research Kit (dVRK).
By aligning the state and action spaces of the physical system with our simulation environment, we successfully transferred the learned dynamics model and policy directly to the real-world setting without modification.

\subsubsection{RL with Safety Controller for No-Go Zone Avoidance}
In this experiment, as shown in \Cref{fig:real_world_needle_gauze}, we evaluate the ability of the system to override a learning-based policy when it attempts to violate safety constraints.
We replicate the \textit{NeedlePick} and \textit{GauzeRetrieve} task in the real world, introducing a sphere no-go zone (radius $r=40.0$ mm) and a cylinder no-go zone (radius $r=34.0$ mm and length $l=30.0$ mm) between the starting position of the robot and the needle.

We compare the standard DEX policy against our SSP-DEX method over 10 trials.
The unconstrained DEX policy breaks into with the no-go zone in 100\% of trials, as it attempted to take the shortest path to the needle.
In contrast, our SSP-DEX policy achieves a 0\% collision rate, successfully deviating from the nominal path to skirt the no-go zone boundary before completing the grasp.

\subsubsection{Multi-Stage Suturing (RL + CLF)}
The suturing experiment, as shown in \Cref{fig:real_world_experiments_suturing_and_lung}, demonstrates the hierarchical switching capability of our framework.
The task is a complete suturing sequence divided into two phases: Grasp Phase (RL) and Insertion Phase (CLF).
During the grasp phase, the robot must locate and grasp a curved suture needle.
This phase uses a RL policy to handle the unstructured nature of the grasp.
Once the needle is grasped successfully, the controller switches to a CLF-based path follower to drive the needle through a suture training board along a pre-planned circular arc.
The transition occurs automatically when the gripper jaw angle indicates a successful grasp.
The CLF controller ($V(s) = \|c(s-s_{des})\|^2$) tracks the reference arc.
This seamless handover highlights the modularity of our method, allowing specialized controllers (RL for dexterity, CLF for precision) to coexist.


\subsubsection{Lung Tumor Resection with Safety Constraints (CLF + CBF)}

In the final experiment, as shown in \Cref{fig:real_world_experiments_suturing_and_lung}, we evaluate the SSP-CLF framework on a mock lung (phantom) tumor resection task.
The objective was to perform a cutting motion along a predefined path while strictly avoiding a no-go zone representing a critical anatomical structure (modeled as a spherical safety region with $r = 90$ mm).

While the CLF formulation generates a reference path for the cutting motion, the CBF imposes a hard constraint to prevent encroachment into the vascular region.
Our results show that without the CBF, the controller strictly follows the reference path, resulting in a violation of the safety region. Conversely, with the safety filter enabled, the controller autonomously deviates from the nominal path to circumnavigate the safety region, smoothly converging back to the reference path once the critical area is passed.
This confirms that our framework strictly prioritizes safety constraints over path following objectives.

\section{Conclusion}
In this work, we presented a unified framework for safe and effective autonomous surgery by integrating Neural ODEs for dynamics modeling, a learning-based method and Control Lyapunov Functions (CLFs) for surgical policy generation and Control Barrier Functions (CBFs) for safety filter.
Specifically, we addressed the challenge of unknown system dynamics by learning a continuous-time model via Neural ODEs, and considered the uncertainty in the learned model.
We defined behavioral constraints and spatial constraints and ensure safety via rigorous application of CBF-based safety filters.
Our unified architecture allows for the seamless integration of high-performance black-box policies (such as policies learned from RL or IL) with strict safety guarantees.
Extensive experiments in both the SurRoL simulator and on a real-world dVRK demonstrated that our method can strictly enforce safety which is critical in surgical environments without significantly compromising task success rates.
By bridging the gap between data-driven generality and model-based safety, this framework provides a robust foundation for the next generation of autonomous surgical assistants.
Future work will focus on incorporating visual inputs to detect no-go zone and define constraints autonomously, which allows real-world deployment with even stronger robustness.




\bibliographystyle{IEEEtran}
\bibliography{ref}
\newpage

\vfill

\end{document}